\pdfoutput=1
\documentclass[3p, times]{elsarticle}
\usepackage{amsmath,amsthm,amssymb,amsfonts}
\allowdisplaybreaks[4]
\usepackage{algorithm, algorithmic}
\usepackage{graphicx}
\usepackage{subfigure}
\usepackage{textcomp}
\usepackage{multirow}
\usepackage{makecell}
\usepackage{booktabs}
\usepackage{url}
\usepackage{color}
\usepackage{setspace}
\usepackage{hyperref}

\newtheorem{theorem}{Theorem}

\hypersetup{
	pdffitwindow=false,
	pdfstartview={FitH},
	colorlinks=true,
	linkcolor=red,
	citecolor=green,
	filecolor=black,
	urlcolor=black
}

\journal{Pattern Recognition}

\begin{document}

\begin{frontmatter}
			
\title{Continuous Conditional Random Field Convolution for Point Cloud Segmentation}
		
\author[1,2]{Fei Yang}
\ead{yangfei92516@163.com}
\author[3]{Franck Davoine}
\ead{franck.davoine@hds.utc.fr}	
\author[2]{Huan Wang}
\ead{wanghuanphd@njust.edu.cn}
\author[1,2]{Zhong Jin\corref{cor}}
\ead{zhongjin@njust.edu.cn}

\cortext[cor]{Corresponding author}
\address[1]{Key Laboratory of Intelligent Perception and Systems for High-Dimensional Information of Ministry of Education, \\\label{key}Nanjing University of Science and Technology, Nanjing 210094, China}
\address[2]{School of Computer Science and Engineering, Nanjing University of Science and Technology, Nanjing 210094, China}
\address[3]{Alliance Sorbonne Universit\'{e}, Universit\'{e} de technologie de Compi\`{e}gne, CNRS, Heudiasyc Lab., CS 60319 - F-60203 Compi\`{e}gne Cedex, France}

\begin{abstract}
Point cloud segmentation is the foundation of 3D environmental perception for modern intelligent systems. To solve this problem and image segmentation, conditional random fields (CRFs) are usually formulated as discrete models in label space to encourage label consistency, which is actually a kind of postprocessing. In this paper, we reconsider the CRF in feature space for point cloud segmentation because it can capture the structure of features well to improve the representation ability of features rather than simply smoothing. Therefore, we first model the point cloud features with a continuous quadratic energy model and formulate its solution process as a message-passing graph convolution, by which it can be easily integrated into a deep network. We theoretically demonstrate that the message passing in the graph convolution is equivalent to the mean-field approximation of a continuous CRF model. Furthermore, we build an encoder-decoder network based on the proposed continuous CRF graph convolution (CRFConv), in which the CRFConv embedded in the decoding layers can restore the details of high-level features that were lost in the encoding stage to enhance the location ability of the network, thereby benefiting segmentation. Analogous to the CRFConv, we show that the classical discrete CRF can also work collaboratively with the proposed network via another graph convolution to further improve the segmentation results. Experiments on various point cloud benchmarks demonstrate the effectiveness and robustness of the proposed method. Compared with the state-of-the-art methods, the proposed method can also achieve competitive segmentation performance.
\end{abstract}


\end{frontmatter}


\section{Introduction}
A conditional random field (CRF) is a kind of probabilistic graphical model (PGM) that is widely employed for structure prediction problems in computer vision. In image segmentation, most previous studies have attempted to model the data affinity in label space with CRFs, where the CRF is formulated as a discrete model. Such a discrete model is usually NP-hard to solve. Nevertheless, some approximate methods have been proposed to solve discrete CRF models efficiently. The most widely used approximate methods in computer vision are $\alpha$-expansion/$\alpha\beta$-swap\cite{Boykov2001fast} and mean-field approximation\cite{Koller2009Probabilistic}. The former is a graph-cut-based method, and the latter belongs to the variational inference-based method. With the rapid development of deep learning techniques, mean-field approximation methods have attracted more attention in the community because the mean-field approximation can be easily embedded in a deep network to build an end-to-end architecture\cite{Arnab2018}. Despite these, discrete CRFs that encourage label consistency in the results can only smooth the object boundaries and remove some small noises. The misclassified object regions in the prediction stage are hard to correct by these kinds of postprocessing CRFs.

Convolutional neural networks (CNNs) have achieved significant success in image, speech, and natural language processing (NLP). However, with the explosion in 3D data such as point clouds, traditional CNNs that are designed for grid-like data can no longer meet all the needs. On the one hand, the point cloud data acquired by 3D sensors are usually large-scale, sparse, and unordered, which is difficult to process with traditional CNNs directly. On the other hand, point cloud segmentation is the foundation of 3D environmental perception in modern intelligent systems, such as intelligent robotics and autonomous driving cars. Therefore, segmenting point clouds with deep learning techniques is an urgent challenge in modern computer vision. Although many deep learning-based methods\cite{qi2017pointnet, li2018pointcnn, thomas2019kpconv} have been proposed to segment point clouds, they focus on the design of the feature extractor to improve the representation ability of the point features. Segmentation consists of two parts: classifying (\textit{what it is}) and locating (\textit{where it is}). For classification tasks, we only need to encode the high-level features, and focus less on where the extracted features should be. The classification result can be inferred from only a single global vector. However, segmentation is a dense prediction task. We need to not only extract the high-level features (\textit{what it is}) but also restore the correspondence between the encoded high-level features and each input point exactly (\textit{where it is}). The restoration process can be understood as solving the problem of \textit{where it is}. In other words, previous studies attempted to make the network understand \textit{what it is} but failed to investigate the problem of \textit{where it is}. Currently, the research on feature decoding is scarce in the literature. To fill this gap, we focus our study on feature decoding rather than feature encoding, considering that the feature encoding methods targeting a classification task are competent for a segmentation task.

To address the problems mentioned above, we reconsider the CRF in feature space to enhance the location ability of the network during feature decoding. Segmentation is a typical structure prediction problem. To address this problem, the CRF is usually formulated as a discrete model to encourage label consistency in the results, which can only smooth the object boundaries and remove some small noises. Inspired by the continuous CRF used for depth completion, we reconsider the CRF as a continuous model in feature space to restore the encoding features for point cloud segmentation. The CRF is enforced to participate in the feature extraction process of the point cloud and thus can fundamentally improve the representation ability of the features rather than just smoothing the labels by postprocessing. Specifically, previous methods usually adopt the kNN (k-nearest neighbors)-based interpolation method to decode high-level features. They roughly place the high-level features at their closest positions during upsampling but ignore the consistency of features in feature space. The details of the high-level features will be lost in such an upsampling process. There is an intuitive assumption that similar points in low-level feature space should also have similar high-level features. Therefore, we employ the CRF model to guide the feature upsampling procedure to restore the details of high-level input point features gradually. In our continuous CRF model, the low-level features in the encoding layers are used as guidance information to guide the high-level feature upsampling. This process has a similar effect to guided filtering, such as joint bilateral filters.

In this paper, we propose a continuous CRF graph convolution (CRFConv) to capture the feature structures during the feature extraction stage. More precisely, we propose using a continuous quadratic energy model to describe the point cloud in feature space, in that a continuous CRF model is mathematically equivalent to a quadric energy model that has a similar form to the diffusion model. Inspired by some diffusion-based graph convolution methods in the community, we formulate the solution process of this model as a message-passing graph convolution, by which the continuous CRF can be embedded in the network for feature extraction. From a CRF perspective, we also prove that the message passing of this graph convolution is equivalent to the mean-field approximation algorithm of a continuous Gaussian CRF model, which indicates that the proposed graph convolution is actually equivalent to a continuous CRF model. After that, we build an encoder-decoder network for point cloud segmentation, in which the decoding modules are carefully designed based on the proposed CRFConv. We use the proposed CRFConv to enhance the location ability (\textit{where it is}) of the network as described above. To enlarge the receptive field, the extracted high-level features in the encoder lose details by successive pooling/downsampling operations. Previous methods only upsample the high-level features with simple linear interpolation in the decoding layers, while our method models the upsampling process with a structure model (i.e.,~the CRFConv), by which the details of the high-level point cloud features can be restored gradually. Furthermore, we show that our continuous CRF can also work collaboratively with the classical discrete CRF to further improve the segmentation results. The discrete CRF is formulated as another message-passing graph convolution by analogy to the CRFConv, by which the discrete CRF can be embedded in our segmentation network to implement an end-to-end dual CRF network. To the best of our knowledge, we are the first to consider CRF models both in feature and label space simultaneously in point cloud segmentation.

Experiments on various datasets, including object-level and scene-level segmentation tasks, demonstrate the effectiveness of the proposed method. The contributions of this paper can be summarized as follows:\footnote{Code is available at \url{https://github.com/yangfei1223/CRFConv}.}
\begin{enumerate}
	\setlength{\itemsep}{0pt}
	\setlength{\parsep}{0pt}
	\setlength{\parskip}{2.5pt}
	\item We propose a continuous CRF graph convolution (CRFConv) to model the data affinity in feature space instead of label space, which enforces the CRFConv to participate in the point cloud feature extraction process. The CRFConv can thus fundamentally improve the representation ability of the features rather than just smoothing the labels by postprocessing.
	\item We design a point cloud segmentation network based on the proposed CRF convolution to enhance the location ability of the network. The CRFConv is embedded in the network decoder to model the upsampling process with a continuous CRF model, by which the details of the high-level point cloud features can be restored gradually during upsampling.
	\item We further implement a dual CRF network by reformulating the classical discrete CRF as another graph convolution. The dual CRF network models the data affinity both in feature space and label space simultaneously to further improve the segmentation results.
	\item We conduct experiments on various challenging point cloud segmentation benchmarks, including synthetic object data and large-scale indoor and outdoor scenes. The experimental results demonstrate the effectiveness and robustness of the proposed method.
\end{enumerate}

The rest of this paper is organized as follows. In section 2, we review the literature and introduce some related basic knowledge about CRF. The continuous CRF graph convolution and theoretical analyses are presented in section 3. In section 4, we present the architecture of the segmentation network in detail and show how to combine the continuous CRF with the classical discrete CRF in our segmentation network. The experimental results, comparisons, and ablation analysis can be found in section 5. Section 6 concludes the work in this paper, in which the limitations and future works are also discussed.

\section{Related Work}
\subsection{Point cloud segmentation}
With the popularity of 3D sensors, 3D data are exploding in human daily life owing to their easy acquisition. One key application of 3D data is point cloud segmentation, which is the foundation of 3D environmental perception. Considering the success of CNNs in image processing, many researchers have attempted to employ deep learning techniques to segment point clouds\cite{Guo2020deep}. In this section, we only review the point cloud segmentation methods based on deep learning. These methods can be divided into four categories: image-based, voxel-based, point-based, and graph-based methods.

\textit{Image-based methods} convert the point cloud as an image-like representation and employ some image-based networks to process the point cloud. For example, Yang et al.\cite{Yang2019fusion} projected LiDAR (light detection and ranging) points to the image plane by the cross-calibration parameters. Instead of converting the LiDAR point cloud to the image plane, Wu et al.\cite{wu2018squeezeseg} projected the LiDAR point cloud to a sphere plane to form a dense and grid-based representation. An alternative solution for 3D point cloud representation is the multiview-based method\cite{kalogerakis20173d}. In other words, the image-based methods convert the 3D point cloud to a 2D grid-like representation to cater to image-based networks. The structural information of the point cloud will inevitably be lost during such conversions.

\textit{Voxel-based methods} borrow ideas from image-based CNNs. In these methods, the point cloud is first split into uniform voxels and then processed by deep 3D networks. Maturana et al. proposed VoxNet\cite{maturana2015voxnet}, an architecture to segment point clouds by integrating a volumetric occupancy grid representation with a supervised 3D CNN. However, the voxel-based presentation for a point cloud is usually time consuming and space wasting. To address this problem, Octree\cite{riegler2017octnet} and Kd-Tree\cite{klokov2017escape}-based approaches have been proposed to save computations by skipping convolution in empty space. Nevertheless, voxel-based representation methods still cause information loss. In addition, it is difficult to balance computations and accuracy. How to determine the voxel resolution for the point cloud, which varies in density, remains a problem.

\textit{Point-based methods} overcome the shortcomings of image-based and voxel-based methods. The unordered and sparse points can be directly processed by point-based methods without transformations. PointNet\cite{qi2017pointnet} was the first work to directly process the points in point clouds with deep learning techniques. Since Qi et al. pioneered this field, many point-based methods have emerged in the community. For example, Qi et al. further extended their PointNet to PointNet++\cite{qi2017pointnet++}, which is a hierarchical version of PointNet. Li et al.\cite{li2018pointcnn} learned a $\chi$-transformation to transform the point cloud to a latent and potential canonical order, by which the typical convolution operator can be applied on the $\chi$-transformed features. \cite{hermosilla2018monte, wu2019pointconv} defined the convolution on point cloud domains using Monte Carlo. Thomas et al.\cite{thomas2019kpconv} proposed the kernel point convolution (KPConv) inspired by the convolution operator on regular domains. Recently, the attention mechanism was also explored to extract the point features for point cloud segmentation in \cite{Feng2020point} and \cite{Zhou2020feature}.

\textit{Graph-based methods} have also been extended to process point clouds with the advent of graph neural networks (GNNs) in recent years. In these methods, they assume that a latent graph exists in the point cloud (e.g.,~kNN or radius graph). Landrieu et al.\cite{landrieu2018large} first presegmented the point cloud to superpoints using an unsupervised cluster algorithm. Then they built a superpoint graph based on the superpoint features extracted by PointNet to segment the point cloud. Wang et al.\cite{wang2019dynamic} proposed EdgeConv, which is dynamically computed in each layer of the network, for high-level tasks on point clouds. DeepGCNs\cite{li2019deepgcns} extended graph convolutional networks (GCNs) to deep models by borrowing concepts from CNNs, specifically residual/dense connections and dilated convolutions. DeepGCNs have also shown their ability on large-scale 3D point cloud segmentation tasks.

Point-based and graph-based methods can both be categorized as geometric deep learning (GDL)\cite{bronstein2017geometric} methods. GDL has shown its advances in processing point clouds. In this paper, the proposed method is also a kind of GDL method. However, previous methods focus on improving the classification ability of the network but ignore its location ability, which is crucial for segmentation. Our work focuses on the decoding rather than the encoding of the point features to enhance the location ability of the network, which is the most different point from existing methods.

\subsection{Conditional random field}
In this section, we first review some basic knowledge about CRF for the convenience of readers who are not familiar with this field. Then, we review the applications of CRFs in computer vision, including but not limited to point cloud processing.

\noindent\textit{Basic knowledge.} Let $\mathcal{G(V,E)}$ be an undirected graph, where $\mathcal{V}$ and $\mathcal{E}$ represent the node and edge sets of the graph, respectively. Denote $x_i$ as a random variable defined on node $i\in\mathcal{V}$. $e(i,j)\in\mathcal{E}$ can be understood as the relationship between the random variables $x_i$ and $x_j$. Thus, $X=\{x_i|i\in\mathcal{V}\}$ can be viewed as a random field on $\mathcal{G}$. If the random variables of $X$ in $\mathcal{G}$ satisfy the Markov property, then we call $X$ a Markov random field (MRF) on $\mathcal{G}$. According to the Hammersley-Clifford theorem, the joint probability distribution of the MRF can be presented with a Gibbs distribution parameterized by the clique potentials defined on $\mathcal{G}$, that is:
\begin{equation}
P(X)=\frac{1}{Z}\prod_{c\in\mathcal{C_G}}\phi_c(X_c),
\label{eq1}
\end{equation} 
where $c$ and $X_c$ denote a maximum clique (complete subgraph) of $\mathcal{G}$ and its associated random variable, respectively. $\mathcal{C_G}$ is the set of all cliques. $\phi_c$ is the clique potential defined on $c$. $Z$ is a partition function to ensure a distribution. Note that $\phi_c$ is constrained to be nonnegative in the Gibbs distribution. In general, we can transform $\phi_c(X_c)$ to its log space. Specifically, we let $\phi_c(X_c)=\exp(-\psi_c(X_c))$, where $\psi_c(X_c)=-\log\phi_c(X_c)$ denotes the energy function defined on $c$. Hence the Gibbs distribution can be simply rewritten as:
\begin{equation}
P(X)=\frac{1}{Z}\exp\big(-E(X)\big),
\label{eq2}
\end{equation} 
where $E(X)=\sum_{c\in\mathcal{C}_G}\psi_c(X_c)$ is the Gibbs energy function defined by all cliques of $\mathcal{G}$. Eq.~(\ref{eq2}) is a log linear model, which ensures a positive distribution. 

The MRF can also be used to describe a conditional distribution. Given another random variable $Y$, if the random variable $X$ on $\mathcal{G}$ is conditional on $Y$, we say the MRF determined by $X$ is a CRF determined by $X$ and $Y$. The conditional distribution $P(X|Y)$ of the CRF can also be represented with a Gibbs distribution, which has the same form as Eq.~(\ref{eq2}). $Y$ and $X$ can be viewed as the observed and latent variables in the CRF, respectively. In other words, the CRF of $X$ and $Y$ can be understood as the MRF of $X$ with an additional input $Y$. For notational convenience, we omit the condition $Y$ in the rest of the paper when we refer to CRF.

\noindent\textit{Applications.} CRFs/MRFs are widely applied to image segmentation for structure prediction as a kind of postprocessing. Kr{\"a}henb{\"u}hl et al.\cite{krahenbuhl2011efficient} proposed the fully connected CRF for image segmentation for the first time. The fully connected CRF also acts as a postprocessing step of a CNN in \cite{chen2017deeplab} to improve the localization accuracy. In addition, Zheng et al.\cite{zheng2015conditional} extended the fully connected CRF as a recurrent neural network (RNN) for object segmentation. Qiu et al\cite{Qiu2020saliency} also integrated the CRF into a CNN to segment saliency objects. Although these methods allow us to jointly train the CNN and the CRF model end-to-end, the CRF still acts as postprocessing in the pipeline. In contrast to the existing approaches that use discrete CRF models, Vemulapalli et al.\cite{vemulapalli2016gaussian} proposed using a Gaussian CRF model for semantic segmentation instead, in which the CRF is still built in the label space to refine the labels. In addition to image segmentation, the application of CRFs/MRFs for point cloud segmentation is also explored. Wu et al. adopt a CRF model as in \cite{chen2017deeplab} to refine the output of their SqueezeNet\cite{wu2018squeezeseg}.

Our continuous CRF model is different from discrete CRFs that usually act as postprocessing to refine the labels. We model the data affinity in feature space with a continuous CRF that essentially participates in the feature extraction process of the point cloud. Hence, our CRF can fundamentally improve the representation ability of the features rather than only smoothing the labels in postprocessing.

\section{The continuous CRF graph convolution }
In this section, we present the CRF graph convolution from a continuous quadratic energy model in the feature space of a point cloud. First, we present how to formulate its solution process as a message-passing algorithm in a graph defined on the point cloud. Then we define the continuous graph convolution based on this message-passing algorithm. Finally, we conduct a theoretical analysis of the proposed graph convolution from a CRF perspective.
\subsection{The continuous quadratic energy model}
Denote $\mathcal{P}=\{(\mathbf{p}_i,\mathbf{z}_i,\mathbf{x}_i)|i<N\}$ as a point cloud with $N$ points. $\mathbf{p}_i\in\mathbb{R}^3$ is the position vector of point $i$ in the 3D Euclidean space. $\mathbf{z}_i\in\mathbb{R}^d$ and $\mathbf{x}_i\in\mathbb{R}^d$ represent the observed and latent features of point $i$, respectively. Assume that the observed feature $\mathbf{z}_i$ is independent of each other. We infer the latent feature $\mathbf{x}_i$ from the observed feature $\mathbf{z}_i$ by considering the affinity in the feature space. Thus, we consider the following quadratic energy function:
\begin{equation}
E(X)=\underbrace{\sum_i(\mathbf{x}_i-\mathbf{z}_i)^\top(\mathbf{x}_i-\mathbf{z}_i)}_{\text{fidelity term}}+\underbrace{\sum_{ij}(\mathbf{x}_i-\mathbf{x}_j)^\top\mathbf{w}_{ij}(\mathbf{x}_i-\mathbf{x}_j)}_{\text{smoothness term}},
\label{eq3}
\end{equation}
where $X=\{\mathbf{x}_i|i<N\}$ denotes the feature set of the latent features. $\mathbf{w}_{ij}\in\mathbb{R}^{d\times d}$ is the weight matrix between features $\mathbf{x}_i$ and $\mathbf{x}_j$, which encodes the relationship between two points. It can be computed as:
\begin{equation}
\mathbf{w}_{ij}=s_{ij}\mathbf{C},
\label{eq4}
\end{equation}
where $s_{ij}\ge 0$ is the scalar similarity between points $i$ and $j$, which can be estimated from the observed feature $\mathbf{z}$.\footnote{$s_{ij}$ is task-relevant and can be learned from data. We specify its form in the next section.} $\mathbf{C}\in\mathbb{R}^{d\times d}\succ\mathbf{0}$ is a compatibility matrix to model the intrarelationship between feature channels and is assumed to be independent of the feature value. There is a similar concept of the compatibility matrix in the classic discrete CRF model. In that case, $\mathbf{C}$ describes the co-occurrence (or compatibility) of different object classes or labels. Since our model is built in feature space, $\mathbf{C}$ actually describes the channel/dimension compatibility of feature space here. In general, we can assume that the channels are independent of each other, and the messages pass in each channel individually in this case. Therefore, it is possible to directly define $\mathbf{C}$ as the identity matrix. Alternatively, it is also possible to automatically learn it from the data to further enhance the expression ability of the model. We cannot parameterize $\mathbf{C}$ directly because of its positive-define constraint. In this paper, we let $\mathbf{C}=\mathbf{c}^\top\mathbf{c}+\epsilon\mathbf{I}$ to ensure positive definition, where $\mathbf{c}$ is a learnable parameter initialized as $\mathbf{c}=\mathbf{I}$. $\epsilon$ is a tiny positive value.

Eq.~(\ref{eq3}) consists of two terms: a fidelity term and a smoothness term. It can be understood that we want the new representation $\mathbf{x}_i$ to be smoother but as close as possible to its original representation $\mathbf{z}_i$. Eq.~(\ref{eq3}) and its variants have been widely applied in the problems of denoising, recovery, and superresolution in computer vision. The latent feature $X$ can be obtained by minimizing Eq.~(\ref{eq3}):
\begin{equation}
X^*=\mathop{argmin}_{X}E(X).
\label{eq5}
\end{equation}

Eq.~(\ref{eq5}) is a quadratic convex optimization problem with a closed-form solution. Denote $\mathbf{Z}=[\mathbf{z}_1^\top, \mathbf{z}_2^\top, ..., \mathbf{z}_N^\top]^\top$ and $\mathbf{X}=[\mathbf{x}_1^\top, \mathbf{x}_2^\top, ..., \mathbf{x}_N^\top]^\top$ as the concatenated feature vector of the point cloud, where $\mathbf{Z}, \mathbf{X}\in\mathbb{R}^{(N\times d)}$. The compact form of Eq.~(\ref{eq3}) can be rewritten as:
\begin{equation}
E(\mathbf{X})=(\mathbf{X-\mathbf{Z}})^\top(\mathbf{X-\mathbf{Z}})+\mathbf{X}^\top(\mathbf{D}-\mathbf{W})\mathbf{X},
\label{eq6}
\end{equation}
where $\mathbf{W}$ is a $N\times N$ block matrix consisting of $\mathbf{w}_{ij}$, which can be computed as $\mathbf{W}(i, j)=\mathbf{w}_{ij}$. $\mathbf{D}$ is a $N\times N$ block diagonal matrix that can be computed as $\mathbf{D}(i,i)=\sum_{j}\mathbf{w}_{ij}$. The derivative of $E(\mathbf{X})$ is calculated with respect to $\mathbf{X}$ and $\partial E(\mathbf{X})/\partial\mathbf{X}=\mathbf{0}$. We can obtain the closed-form solution of $E(\mathbf{X})$:
\begin{equation}
\mathbf{X}^*=(\mathbf{I}+\mathbf{D}-\mathbf{W})^{-1}\mathbf{Z},
\label{eq7}
\end{equation}
where $\mathbf{I}+\mathbf{D}-\mathbf{W}$ is a $(N\times d)\times(N\times d)$ matrix. 

To solve the linear system, we need to calculate the inverse of such a large matrix, which is computationally expensive, especially when $N$ and $d$ are both large. Alternatively, we adopt an iterative strategy instead of using the closed-form solution. Recalling Eq.~(\ref{eq3}), we calculate the partial derivation of $E(X)$ with respect to $\mathbf{x}_i$ and let $\partial E(X)/\partial \mathbf{x}_i=\mathbf{0}$. Then, we can obtain an iterative solution for $\mathbf{x}_i$ under the coordinate descent framework:
\begin{equation}
\mathbf{x}_i^*=(\mathbf{I}+\sum_{j}\mathbf{w}_{ij})^{-1}(\mathbf{z}_i+\sum_{j}\mathbf{w}_{ij}\mathbf{x}_j).
\label{eq8}
\end{equation}
Since Eq.~(\ref{eq3}) is convex, the coordinate descent algorithm can guarantee the global minimum, which is validated in the experimental section. If we further assume that point $i$ is only related to its neighbor points $\mathcal{N}(i)=\big\{j\big|\lVert\mathbf{p}_i-\mathbf{p}_j\rVert_2^2\le r\big\}$ in the point cloud and normalize the similarity to $\hat{s}_{ij} = s_{ij}/\sum_{j\in\mathcal{N}(i)}s_{ij}$, Eq.~(\ref{eq8}) can be further simplified to:
\begin{equation}
\mathbf{x}_i^*={(\mathbf{I}+\mathbf{C})}^{-1}(\mathbf{z}_i+\mathbf{C}\sum_{j\in\mathcal{N}(i)}\hat{s}_{ij}\mathbf{x}_j).
\label{eq9}
\end{equation}

It is easy to find that Eq.~(\ref{eq3}) actually contains a graph structure implicitly. We denote the graph defined on a point cloud as $\mathcal{G}_\mathcal{P}(\mathcal{V}_\mathcal{P},\mathcal{E}_\mathcal{P})$, where $\mathcal{V}_\mathcal{P}$ is the node set that contains all points ($|\mathcal{V}_\mathcal{P}|=N$). $\mathcal{E}_\mathcal{P}$ represents the edge set that indicates whether two points are related to each other. Therefore, the update equation Eq.~(\ref{eq9}) can be described as a message-passing process in $\mathcal{G}_\mathcal{P}$, as shown in Algorithm \ref{alg1}.

\begin{algorithm}
	\caption{The message-passing process of Eq.~(\ref{eq3})}
	\label{alg1}
	\begin{algorithmic}[1]
		\REQUIRE Point cloud graph $\mathcal{G}_\mathcal{P}(\mathcal{V}_\mathcal{P},\mathcal{E}_\mathcal{P})$, observed feature $Z=\{\mathbf{z}_i|i\in\mathcal{V}_\mathcal{P}\}$, maximum iteration steps $T$.
		\STATE Initialize: $\mathbf{x}_i\leftarrow\mathbf{z}_i$ for $i\in\mathcal{V}_\mathcal{P}$
		\STATE Compute the normalized similarity: $\hat{s}_{ij}$ for $e(i,j)\in\mathcal{E}_\mathcal{P}$
		\FOR{$t=1: T$}
		\FOR{$i\in\mathcal{V}_\mathcal{P}$} 
		\STATE Message passing: $msg_{j\to i}=\sum_{e(i,j)\in\mathcal{E}_\mathcal{P}}\hat{s}_{ij}\mathbf{x}_j$
		\STATE Compatibility transformation: $msg'_{j\to i}=\mathbf{C}msg_{j\to i}$
		\STATE Add unary: $\mathbf{x'}_i=\mathbf{z}_i+msg'_{j\to i}$
		\STATE Normalize: {$\mathbf{x}_i={(\mathbf{I}+\mathbf{C})}^{-1}\mathbf{x'}_i$}
		\ENDFOR
		\ENDFOR
		\ENSURE Latent feature $X=\{\mathbf{x}_i|i\in\mathcal{V}_\mathcal{P}\}$.
	\end{algorithmic}
\end{algorithm}

\subsection{The message-passing graph convolution}
Gilmer et al. proposed a universal GNN paradigm named message-passing neural networks (MPNNs) in \cite{gilmer2017neural}. The forward pass of MPNN has two phases: message passing and readout. In fact, most graph convolutional networks previously proposed\cite{kipf2017semi,velivckovic2018graph,monti2017geometric} follow such a message-passing paradigm. Accordingly, we can also reformulate Algorithm \ref{alg1} as a message-passing graph convolution. 

\begin{figure}[!h]
	\centerline{\includegraphics[width=0.5\columnwidth]{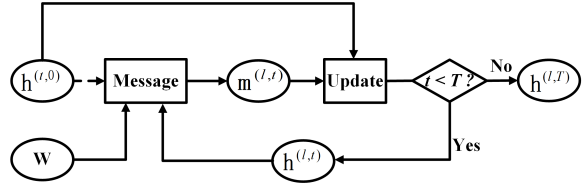}}
	\caption{The data flow of the message-passing module of the proposed graph convolution.} 
	\label{fig1}
\end{figure}

Denote $\mathbf{x}_i^{l}$ as the feature vector of node $i$ in the $l_\text{th}$ layer of the network. Our message-passing graph convolution can be defined as:

\noindent\textit{Initial step}:
\begin{equation}
\mathbf{h}_i^{(l, 0)}=\mathbf{z}_i^{(l)}=\text{MLP}(\mathbf{x}_i^{(l-1)}),
\label{eq10}
\end{equation}
\noindent\textit{Message-passing step}:
\begin{equation}
\text{Message:\quad}\mathbf{m}_i^{(l,t)} = \mathbf{C}\sum_{j\in\mathcal{N}(i)}\hat{s}_{ij}\mathbf{h}_j^{(l,t-1)},
\label{eq11}
\end{equation}
\begin{equation}
\text{Update:\quad}\mathbf{h}_i^{(l,t)} = {(\mathbf{I}+\mathbf{C})^{-1}}(\mathbf{h}_i^{(l,0)}+\mathbf{m}_i^{(l,t)}),
\label{eq12}
\end{equation}
\noindent\textit{Readout step}:
\begin{equation}
\mathbf{x}_i^{l} = \sigma(\mathbf{h}_i^{(l, T)}),
\label{eq13}
\end{equation}
where $\text{MLP}(\cdot)$ denotes the multilayer perceptron which is used for the pointwise feature transformation. $\mathbf{m}_i^{(l,t)}$ and $\mathbf{h}_i^{(l, t)}$ denote the message and the hidden state of node $i$ in the $l_{th}$ layer after $t$ message-passing steps, respectively. $\sigma(\cdot)$ acts as a nonlinear activation function in the read out step. The data flow of the message-passing step is illustrated in Figure \ref{fig1}. Unlike most one-step graph convolution methods, the message-passing phase of our graph convolution, which consists of a message function Eq.~(\ref{eq11}) and an update function Eq.~(\ref{eq12}), runs $T$ times.

\subsection{The CRF Perspective}
Consider a Gibbs distribution in Eq.~(\ref{eq2}) with the energy function given in Eq.~(\ref{eq3}). According to the Hammersley-Clifford theorem, in this case, Eq.~(\ref{eq2}) actually determines a continuous Gaussian CRF model defined on $\mathcal{G}_\mathcal{P}(\mathcal{V}_\mathcal{P},\mathcal{E}_\mathcal{P})$. Considering CRF, the maximization of $P(X)$ is also equivalent to the minimization of the energy function $E(X)$.
\begin{theorem}
The coordinate decent of Eq.~(\ref{eq3}) is equivalent to the mean-field approximation of Eq.~(\ref{eq2}) when the approximate distribution $Q_i\sim N(\mathbf{\mu}_i,\mathbf{\Sigma}_i)$.
\end{theorem}
\noindent\textit{Proof}. The goal of the mean-field approximation algorithm is to find a distribution $Q(X)$ that can be factorized as $Q(X)=\prod_{i}Q_i(\mathbf{x}_i)$ to approximate the original distribution $P(X)$, where we have $Q_i(\mathbf{x}_i)\sim N(\mathbf{x}_i;\mathbf{\mu}_i,\mathbf{\Sigma}_i)$. Then, we can obtain $Q(X)$ by minimizing the Kullback-Leibler divergence $KL(Q||P)$ between $Q(X)$ and $P(X)$.
We obtain $\{(\mathbf{\mu}_i^*,\mathbf{\Sigma}_i^*)|i\in\mathcal{V}_\mathcal{P}\}$ by solving the following problem:
\begin{equation}
\begin{aligned}
\{(\mathbf{\mu}_i^*,\mathbf{\Sigma}_i^*)|i\in\mathcal{V}_\mathcal{P}\}
&=\mathop{argmin}_{\{(\mathbf{\mu}_i,\mathbf{\Sigma}_i)|i\in\mathcal{V}_\mathcal{P}\}}KL(Q||P)\\
&=\mathop{argmin}_{\{(\mathbf{\mu}_i,\mathbf{\Sigma}_i)|i\in\mathcal{V}_\mathcal{P}\}}F[Q(X;\{(\mathbf{\mu}_i,\mathbf{\Sigma}_i)|i\in\mathcal{V}_\mathcal{P}\}),E(X)],
\end{aligned}
\label{eq15}
\end{equation}
where $F[\cdot]$ is an energy functional about the energy function $E(X)$ and the approximation distribution $Q(X)$. Since the forms of $E(X)$ and $Q(X)$ are determined in our case, $F$ can be viewed as a function of $\{(\mathbf{\mu}_i,\mathbf{\Sigma}_i)|i\in\mathcal{V}_\mathcal{P}\}$ directly. $Q_i$ is supposed to be independent and identically distributed. Without losing generality, we consider each $\mathbf{\mu}_i$ and $\mathbf{\Sigma}_i$ individually. By setting $\partial F/\partial \mathbf{\mu}_i=0$ and $\partial F/\partial \mathbf{\Sigma}_i=0$, we can obtain the update equations of $\mathbf{\mu}_i$ and $\mathbf{\Sigma}_i$, respectively:
\begin{equation}
\mathbf{\mu}_i^*=(\mathbf{I}+\sum_{j\in\mathcal{N}(i)}\mathbf{w}_{ij})^{-1}(\mathbf{z}_i+\sum_{j\in\mathcal{N}(i)}\mathbf{w}_{ij}\mathbf{\mu}_j).
\label{eq16}
\end{equation}
\begin{equation}
\mathbf{\Sigma}_i^*=\frac{1}{2}(\mathbf{I}+\sum_{j\in\mathcal{N}(i)}\mathbf{w}_{ij})^{-1}.
\label{eq17}
\end{equation}

Actually, we are only interested in the mean vector $\mathbf{\mu}_i$ of $Q_i$ in that $max\big(Q_i(\mathbf{x}_i)\big)=Q_i(\mathbf{\mu}_i)$. We find that the update equation of $\mathbf{\mu}_i$ in the mean-field algorithm is equivalent to the update equation of $\mathbf{x}_i$ in the coordinate descent algorithm. In other words, the coordinate descent of Eq.~(\ref{eq3}) is equivalent to the mean-field approximation of a continuous CRF (Eq.~(\ref{eq2})), by which Theorem 1 can be proved. Hence, we name the proposed message-passing graph convolution continuous CRF graph convolution (CRFConv). For clarity reasons, we only provide a brief proof here. The detailed proof of Theorem 1 can be found in the supplemental material.

\section{The point cloud segmentation network}
\begin{figure*}[!h]
	\centerline{\includegraphics[width=0.8\columnwidth]{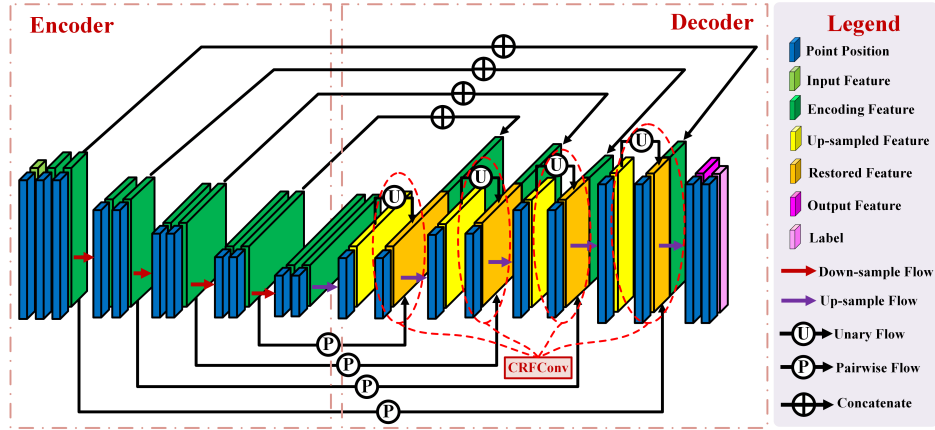}}
	\caption{Overview of the proposed point cloud segmentation network. The red dotted circles indicate where the CRFConv was embedded.}
	\label{fig2}
\end{figure*}

In this section, we show how to utilize the proposed CRFConv for point cloud segmentation. We design a point cloud segmentation network that includes two parts: an encoder and a decoder. An overview of the network architecture is shown in Figure \ref{fig2}. The CRFConv is embedded in the decoder to decode the high-level features. Specifically, we model the upsampling process with the CRF model rather than linear interpolation, by which the lost details of high-level features caused by the encoder can be restored gradually during feature decoding.

\subsection{The encoder}
Because point clouds are unordered, sparse, and continuous in 3D Euclidean space, the convolution on a point cloud is different from that on an image. Analogous to image convolution, we adopt the convolution defined on irregular domains to extract the point features in the encoder. The standard convolution operation is defined as:
\begin{equation}
f*g(\mathbf{x})=\int_{-\infty}^{+\infty}f(\mathbf{y})g(\mathbf{x-y})d\mathbf{y}=\sum_{\mathbf{y}\in\mathcal{N}(\mathbf{x})}f(\mathbf{y})g(\mathbf{x-y}),
\label{eq21}
\end{equation}
where $f$ is the signal to be convolved, and $g$ is the convolution kernel. In a regular domain (e.g.,~image), the offset $(\mathbf{x-y})$ is fixed with respect to different kernel centers, which allows us to directly parameterize $g$ with shared parameters. However, for an irregular domain (e.g.~,point cloud), the offset $(\mathbf{x-y})$ and the neighborhood size $|\mathcal{N}(\mathbf{x})|$ may vary at different positions, which leads to the fact that we cannot parameterize $g$ directly. In the literature, researchers have attempted different methods to define the convolution on irregular domains\cite{li2018pointcnn, thomas2019kpconv,  wang2018deep, hermosilla2018monte, wu2019pointconv}. In this paper, we follow \cite{wang2018deep, hermosilla2018monte, wu2019pointconv} and parameterize the convolution kernel $g$ as:
\begin{equation}
g(\mathbf{x-y})=\text{MLP}(\mathbf{x-y};\mathbf{\Theta}),
\label{eq22}
\end{equation}
where $\mathbf{\Theta}$ is the learnable weights of the MLP. Therefore, the point convolution (PointConv) used in this paper can be written as:
\begin{equation}
f*g(\mathbf{x})=\frac{1}{|\mathcal{N}(\mathbf{x})|}\sum_{\mathbf{y}\in{\mathcal{N}(\mathbf{x})}}f(\mathbf{y})\text{MLP}(\mathbf{x-y};\mathbf{\Theta}),
\label{eq23}
\end{equation}
in which normalization is necessary to make it robust to various neighbor sizes.

\begin{figure}[!h]
	\centerline{\includegraphics[width=0.5\columnwidth]{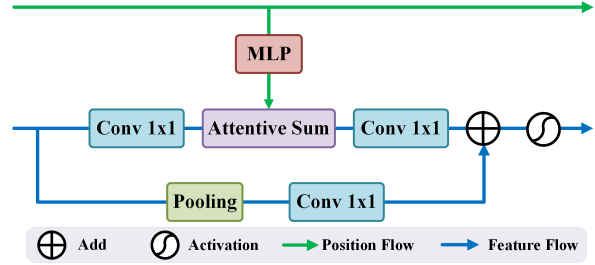}}
	\caption{The PointConv used in the encoding layers. Note that in the residual path, pooling and $1\times 1$ convolution are adopted to align the point cloud size or feature dimension if necessary.} 
	\label{fig3}
\end{figure}

$d_\text{in}$ and $d_\text{out}$ are denoted as the input and output feature dimensions, respectively. For each PointConv, the kernel $g(\cdot)$ is evaluated $N\times|\mathcal{N}|\times d_\text{in}\times d_\text{out}$ times. This is computationally expensive in practice, especially when both the number of points and the feature dimension are large. Inspired by the depthwise separable convolution of image networks, we separate canonical PointConv into a depthwise point convolution and several $1\times 1$ convolutions. Figure \ref{fig3} illustrates our PointConv layer in detail. Specifically, we first use a $1\times 1$ convolution to reduce the dimension of the input feature from $d_\text{in}$ to $d_\text{in}/r$ with a proportion of $r$. Then, we evaluate $N\times|\mathcal{N}|\times(d_\text{in}/r)$ kernels for depthwise point convolution. Finally, another $1\times 1$ convolution with $(d_\text{in}/r)\times d_\text{out}$ parameters is adopted to produce the output features. Moreover, batch normalization (BN) and residual connections are also adopted for PointConv. Dilated kNN neighbors are constructed for dilated PointConv to increase the receptive field of the encoder, which is proven to be helpful in image segmentation\cite{chen2017deeplab}.

\subsection{The decoder}
In the literature, successive pooling operations or strided convolutions are adopted to increase the receptive field of the encoder. In the decoding stage, they simply adopt linear interpolation to upsample the high-level features, by which the details of these features are lost. Such linear interpolation only considers the positions of the features but ignores their continuity in feature space. The high-level features become increasingly coarser during decoding. Previous methods skip connections to relieve this situation. However, no previous study has fundamentally attempted to solve this problem. To address this problem, we propose using the CRFConv to restore the details of the high-level features during decoding. In our method, not only the point positions but also the distribution of the low-level features are considered for feature decoding. CRFConv is utilized to model the data affinity in feature space to generate high-resolution and detailed high-level features.

\begin{figure}[!h]
	\centering 
	\subfigure[The continuous CRF convolution layer.] 
	{ 
		\label{fig4a} 
		\includegraphics[width=0.5\columnwidth]{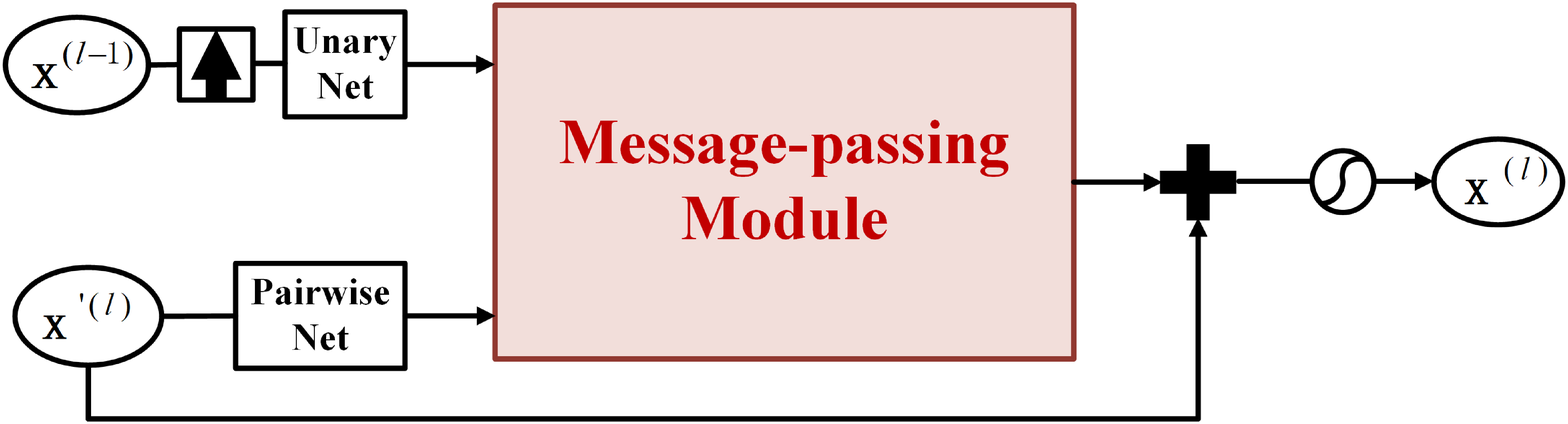} 
	} 
	\subfigure[The discrete CRF convolution layer.] 
	{ 
		\label{fig4b} 
		\includegraphics[width=0.45\columnwidth]{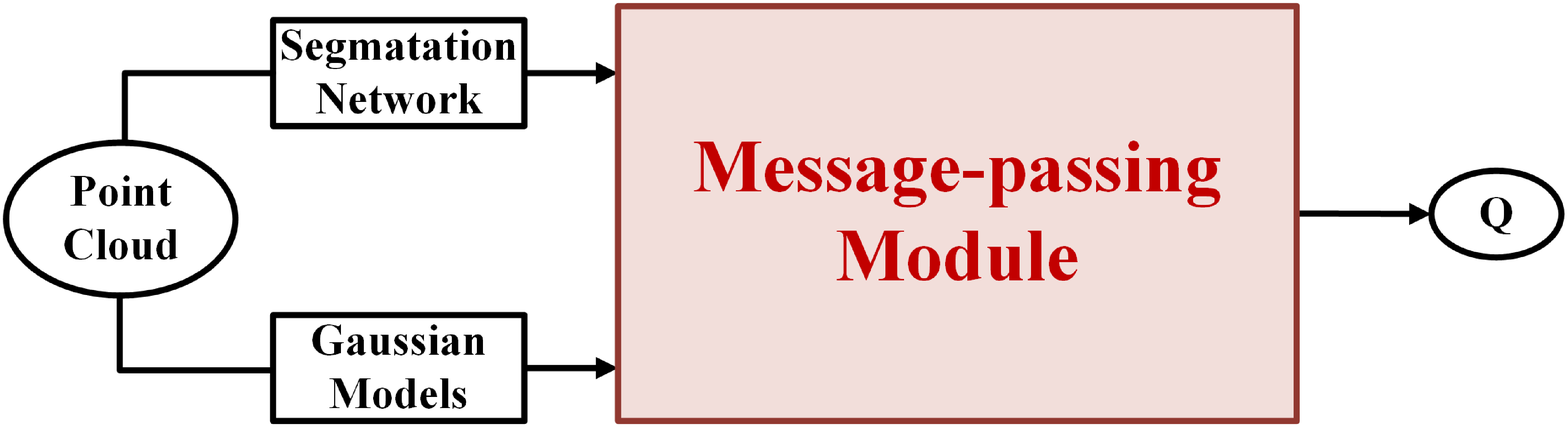} 
	} 
	\caption{The continuous and discrete CRF convolution layer used in the point cloud segmentation network. The data flow of the message-passing modules here can be represented with Figure \ref{fig1}.} 
	\label{fig4} 
\end{figure}

Denote $\mathbf{x}^{(l-1)}$ and $\mathbf{x}^{l}$ as the high-level feature to be upsampled (the input feature in the decoder) and the upsampled high-level feature (the output feature in the decoder), respectively. Let $\mathbf{x'}^{(l)}$ be the corresponding lower-level feature in the encoder that has the same resolution as $\mathbf{x}^{(l)}$. Figure \ref{fig4a} gives the architecture of our decoding layer. The high-level feature is initially upsampled by kNN interpolation. Then, a \textit{unary net} is immediately adopted to extract the unary feature, which produces $\mathbf{z}^{(l)}$ as described in Eq.(\ref{eq10}). Then, we dynamically compute the similarity $s_{ij}$. We adopt a \textit{pairwise net} that takes the corresponding feature $\mathbf{x'}^{(l)}$ in the encoding layer to compute $s_{ij}$. The CRFConv takes these two inputs to generate the restored feature via the message-passing module, as shown in Figure \ref{fig1}. Once the message-passing completes, the restored feature is further concatenated with the corresponding lower-level feature $\mathbf{x'}^{(l)}$ in the encoding layer to feed the next decoding layer after activation.
\paragraph{Unary Net}
We simply take an MLP as the unary net, which conducts a pointwise feature transformation for the point features. The unary net plays the roles of dimension reduction and feature fusion in the decoding layers.
\paragraph{Pairwise Net}
The pairwise net is used to compute the similarity $s_{ij}$ of each point pair $(i,j)$. For each high-level point feature pair $(\mathbf{x}_i, \mathbf{x}_j)$, we compute its similarity with the corresponding point feature pair $(\mathbf{x}_i^\prime, \mathbf{x}_j^\prime)$ in a lower-level feature space. We adopt the Mahalanobis distance in the low-level feature space to compute the similarity $s_{ij}$. The vanilla Mahalanobis distance is computed as:
\begin{equation}
d^2(\mathbf{x}_i^\prime,\mathbf{x}_j^\prime)={(\mathbf{x}_i^\prime-\mathbf{x}_j^\prime)}^\top\mathbf{M}(\mathbf{x}_i^\prime-\mathbf{x}_j^\prime)=(\mathbf{x}_i^\prime-\mathbf{x}_j^\prime)^\top\mathbf{P}\mathbf{P}^\top(\mathbf{x}_i^\prime-\mathbf{x}_j^\prime)=\lVert\mathbf{P}^\top\mathbf{x}_i^\prime-\mathbf{P}^\top\mathbf{x}_j^\prime\rVert_2^2,
\label{eq24}
\end{equation}
where $\mathbf{x}_i^\prime,\mathbf{x}_j^\prime\in\mathbb{R}^d$, $\mathbf{M}\in\mathbb{R}^{d\times d}\succ\mathbf{0}$, $\mathbf{P}\in\mathbb{R}^{d\times d'}$. The Mahalanobis distance is actually the Euclidean distance on a projected space determined by the linear projection matrix $\mathbf{P}$, which may fail to model nonlinear data effectively. In this paper, we reformulate $\mathbf{P}$ as a nonlinear projection to better model nonlinear data. We let the network learn a measure function automatically with an MLP. Thus, the improved Mahlanobis distance used in this paper can be computed as:
\begin{equation}
d^{\prime 2}(\mathbf{x}_i^\prime,\mathbf{x}_j^\prime)=\lVert \text{MLP}(\mathbf{x}_i^\prime)-\text{MLP}(\mathbf{x}_j^\prime) \rVert_2^2.
\label{eq25}
\end{equation}
Accordingly, the normalized similarity $\hat{s}_{ij}$ is given as:
\begin{equation}
\hat{s}_{ij}=\frac{\exp(-d^{\prime 2}(\mathbf{x}_i^\prime,\mathbf{x}_j^\prime))}{\sum_{j\in\mathcal{N}(i)}\exp(-d^{\prime 2}(\mathbf{x}_i^\prime,\mathbf{x}_j^\prime))},
\label{eq26}
\end{equation}
which can be easily implemented with a softmax layer in the network.

\subsection{Classical discrete CRF as graph convolution}
Most previous studies utilize the discrete CRF model to refine the labels in segmentation tasks, while our CRF is a continuous model defined in feature space. Hence our continuous CRF can work collaboratively with the classical discrete CRF to further improve the segmentation performance. In image segmentation, the energy function of the discrete CRF is usually defined as a pairwise model\cite{krahenbuhl2011efficient}:
\begin{equation}
E(X)=\sum_i\psi_\text{u}(x_i)+\sum_i\sum_{j\in\mathcal{N}(i)}\psi_\text{p}(x_i,x_j),
\label{eq27}
\end{equation}
where $x_i\in\mathcal{L}$ presents the label of pixel (node) $i$. Here $\mathcal{L}$ is a discrete label set. $\psi_\text{u}(x_i)\ge0$ is the unary potential, which gives a penalty when each pixel is assigned a label. In the postprocessing CRFs, the unary potential can be simply computed as $\psi_\text{u}(x_i)=-\log{p(x_i)}$, where $p(x_i)$ is the probability predicted by a discriminative classifier. $\psi_\text{p}(x_i,x_j)=\mu(x_i,x_j)K(\mathbf{f}_i,\mathbf{f}_j)$ is the pairwise potential, which penalizes two connected pixels with different labels. Here, $\mu(\cdot, \cdot)$ measures the compatibility between different classes. $K(\cdot,\cdot):\mathbb{R}^d\times\mathbb{R}^d\to\mathbb{R^+}$ is a positive function defined on a feature space of the pixels. The update equation of the approximate distribution $Q_i$ in the mean-field approximation of Eq.~(\ref{eq27}) is given in \cite{krahenbuhl2011efficient}:
\begin{equation}
Q_i^*(x_i)=\frac{1}{Z_i}\exp\bigg\{-\psi_\text{u}(x_i)-\sum_{l\in\mathcal{L}}\mu(x_i,l)\sum_{j\in\mathcal{N}(i)}K(\mathbf{f}_i,\mathbf{f}_j)Q_j(l)\bigg\}.
\label{eq28}
\end{equation}

We show that Eq.~(\ref{eq27}) can also be used for point cloud segmentation. For each point $i$, let $\mathbf{p}_i$ be the output probability vector of the segmentation network and $\mathbf{q}_i$ be the vector version of $Q_i$. Eq.~(\ref{eq28}) can be compactly rewritten as:
\begin{equation}
\mathbf{q}_i^*=\text{Softmax}\bigg(\log(\mathbf{p}_i)-\mathbf{C}\sum_{j\in\mathcal{N}(i)}w_{ij}\mathbf{q}_j\bigg),
\label{eq29}
\end{equation}
where we denote $w_{ij}=K(\mathbf{f}_i,\mathbf{f}_j)$. $\mathbf{C}\in\mathbb{R}^{|\mathcal{L}|\times|\mathcal{L}|}$ is a matrix that encodes the compatibility of labels. Here, we have no constraint on $\mathbf{C}$ and learn it from the data. If we set $\mathbf{C}=\mathbf{I}$, it will degenerate to the Potts model, that is, $\mu(x_i,x_j)=0\ \text{if}\ x_i=x_j\ \text{otherwise}\ 1$, which is widely adopted in most previous image segmentation CRFs.

\begin{algorithm}
	\caption{Message passing of the discrete CRF}
	\label{alg2}
	\begin{algorithmic}[1]
		\REQUIRE Point cloud graph $\mathcal{G}_\mathcal{P}=(\mathcal{V}_\mathcal{P},\mathcal{E}_\mathcal{P})$, initial distribution $P=\{\mathbf{p}_i|i\in\mathcal{V}_\mathcal{P}\}$, maximum iteration step $T$.
		\STATE Initialize: $\mathbf{q}_i\leftarrow\mathbf{p}_i$ \textbf{for} $i\in\mathcal{V}_\mathcal{P}$
		\STATE Compute weights: $w_{ij}=K(\mathbf{f}_i,\mathbf{f}_j)$ \textbf{for} $e(i,j)\in\mathcal{E}_\mathcal{P}$
		\FOR{$t=1: T$}
		\FOR{$i\in\mathcal{V}_\mathcal{P}$}
		\STATE Message-passing: $msg_{j\to i}=\sum_{e(i,j)\in\mathcal{E}_\mathcal{P}}w_{ij}\mathbf{q}_j$
		\STATE Compatibility transformation: $msg^\prime_{j\to i}=\mathbf{C}msg_{j\to i}$
		\STATE Add unary: $\widetilde{\mathbf{q}}_i=\log(\mathbf{p}_i)-msg^\prime_{j\to i}$
		\STATE Normalize: $\mathbf{q}_i=\text{Softmax}(\widetilde{\mathbf{q}}_i)$ 
		\ENDFOR
		\ENDFOR
		\ENSURE Approximate distribution $Q=\{\mathbf{q}_i|i\in\mathcal{V}_\mathcal{P}\}$.
	\end{algorithmic}
\end{algorithm}
Eq.~(\ref{eq29}) can be formulated as another message-passing algorithm on the point cloud graph $\mathcal{G}_\mathcal{P}(\mathcal{V}_\mathcal{P},\mathcal{E}_\mathcal{P})$, as shown in Algorithm \ref{alg2}. Analogous to CRFConv, we can further reformulate Algorithm \ref{alg2} as another message-passing graph convolution for point cloud segmentation:

\noindent\textit{Initial step}:
\begin{equation}
\mathbf{h}_i^{(0)} = \mathbf{p}_i,
\label{eq30}
\end{equation}
\noindent\textit{Message-passing step}:
\begin{equation}
\text{Message:\quad}\mathbf{m}_i^t = \sum_{j\in\mathcal{N}(i)}w_{ij}\mathbf{h}_j^{(t-1)},
\label{eq31}
\end{equation}
\begin{equation}
\text{Update:\quad}\mathbf{h}_i^t=\text{Softmax}\bigg(\log(\mathbf{p}_i)-\mathbf{C}\mathbf{m}_i^t\bigg),
\label{eq32}
\end{equation}
\noindent\textit{Readout step}:
\begin{equation}
\mathbf{q}_i=\mathbf{h}_i^T.
\label{eq33}
\end{equation}

Following \cite{krahenbuhl2011efficient}, we set $w_{ij}=\sum_{m=1}^M\omega_mk_m(\mathbf{f}_i,\mathbf{f}_j)$ as the linear combination of multiple Gaussian, where $k_m(\mathbf{f}_i,\mathbf{f}_j)=\exp\big(-(\mathbf{f}_i-\mathbf{f}_j)^\top\mathbf{\Sigma}_m^{-1}(\mathbf{f}_i-\mathbf{f}_j)\big)$. Instead of adopting artificially designed parameters such as \cite{krahenbuhl2011efficient}, we learn $\omega_m$ and $\mathbf{\Sigma}_m$ of each Gaussian component from the data automatically. To ensure positive definition and avoid the inverse operation, we let $\mathbf{\Sigma}_m^{-1}=\mathbf{P}\mathbf{P}^\top$, where $\mathbf{P}\in\mathbb{R}^{d\times d'}$,  which can be implemented with a linear layer having $d'$ outputs. Similarly, the weight sum operation can also be implemented with another linear layer having only one output. The data flow of the discrete CRF convolution layer is shown in Figure \ref{fig4b}. It can be added at the end of our segmentation network as an additional graph convolution layer to form a dual CRF network. In addition, the whole network can be trained end-to-end.

\section{Experiments}
The experimental platform is a PC equipped with an Intel Core i7 9700K CPU, 16 GB RAM, and an Nvidia TITAN RTX GPU (24 GB). The algorithm is implemented with \textit{PyTorch}\cite{paszke2019pytorch} and \textit{PyTorch Geometric}\cite{fey2019fast} on Ubuntu 16.04.

\subsection{Datasets}
The experiments are conducted on various 3D point cloud segmentation datasets, including ShapeNet\cite{yi2016scalable} for object-level evaluation and S3DIS\cite{armeni20163d} and Semantic3D\cite{hackel2017semantic3d} for indoor and outdoor scene-level evaluation.

\paragraph{ShapeNet} 
ShapeNet is a 3D object model dataset that contains 16,880 object models from 16 categories. Each 3D object consists of approximately 2k points. ShapeNet Parts is the part segmentation benchmark for the ShapeNet models, in which each model annotates 2 to 6 parts with a part label (50 parts in total).
\paragraph{S3DIS} 
S3DIS is a large-scale indoor 3D scene segmentation dataset that was collected on six large-scale indoor areas from 3 different buildings. The scene is annotated with 13 categories (including clutter). Following \cite{qi2017pointnet}, we organize the dataset by rooms (271 rooms in total). In our experiments, we split the training and test sets according to area.
\paragraph{Semantic3D}
Semantic3D is a large-scale outdoor point cloud segmentation online benchmark. This benchmark provides a large labeled 3D point cloud dataset of natural scenes with over 4 billion points in total, which covers a range of diverse urban scenes. We conduct our experiments on the reduced-8 benchmark, which contains 15 scenes for training and 4 reduced scenes for testing with 8 semantic categories.

\subsection{Experimental Settings}
The encoder of our segmentation network contains five convolution blocks, each of which consists of two successive PointConv layers. The cloud resolution is reduced by the farthest point sampling (fps) with a ratio of $r$ after each convolution block; additionally, the feature dimension $d$ is doubled. In our experiment, we set $r = [1.0, 0.25,\ 0.375,\ 0.375,\ 0.375]$ and $d=[64,\ 128,\ 256,\ 512,\ 1024]$. The decoder consists of four corresponding CRFConv layers to restore the cloud resolution to the original input size gradually. We use kNN to search the neighbors for each point. The kernel (neighbor) size $k$ is set to 32 in the first layer and 16 in the remaining layers. The dilated kNN is only available in the encoder. We set the dilation rate $dil=[1,\ 2,\ 4,\ 4,\ 2]$ in the five PointConv blocks. Two fully connected layers follow the last decoding layer to obtain per-point labels. The discrete CRF is added after the last fully connected layer if necessary. To make fair comparisons with the previous methods, the discrete CRF convolution is not adopted in the comparison experiments. The performance of the discrete CRF convolution is evaluated in the ablation study. The detailed network architecture and more experimental details can be found in the supplementary material.

\subsection{Object-Level Evaluation}
We conduct a shape part segmentation experiment on the ShapeNet Parts benchmark. Instead of training each category individually, we train a model with all part labels (50 in total). The normalized $x$, $y$, $z$ position and the normal vector $n_x$, $n_y$, $n_z$ of each point are taken as the input features. Because our network supports various input point cloud sizes, we do not sample the points as in previous methods\cite{qi2017pointnet, qi2017pointnet++, li2018pointcnn}. The intersection over union (IoU) of the object parts is used as the metric to evaluate the performance. The instance part IoU (pIoU) and mean class pIoU (mpIoU) are given in Table \ref{tab2}. Compared with nearly twenty current popular methods, the performance of the proposed method exceeds most previous methods. From Table \ref{tab2}, we can find that KPConv\cite{thomas2019kpconv} performs best overall, which may be due to the better feature encoding ability. Even compared with state-of-the-art methods, the proposed method can achieve competitive performance. Please see the supplementary material for some visualization results.

\begin{table}[!htbp]
	\caption{Performance of the proposed method on the ShapeNet Part segmentation benchmark.}
	\footnotesize
	\renewcommand\arraystretch{1.0}
	\setlength{\tabcolsep}{0.5mm}
	\centering
	\resizebox{\textwidth}{!}
	{
	\begin{tabular}{c|ccccccccccccccccccc}
		\toprule
		&Methods &\textbf{pIoU} &\textbf{mpIoU} &air &bag &cap &car &chair &ear &guitar &knife &lamp &laptap &motor &mug &pistol &rocket &bord &tabel \\
		\midrule
		\multirow{5}*{2017} &Kd-Net\cite{klokov2017escape} &82.3 &77.4 &80.1 &74.6 &74.3 &70.3 &88.6 &73.5 &90.2 &87.2 &81.0 &94.9 &57.4 &86.7 &78.1 &51.8 &69.9 &80.3 \\
		&PointNet\cite{qi2017pointnet} &83.7 &80.4 &83.4 &78.7 &82.5 &74.9 &89.6 &73.0 &91.5 &85.9 &80.8 &95.3 &65.2 &93.0 &81.2 &57.9 &72.8 &80.6 \\
		&PointNet++\cite{qi2017pointnet++} &85.1 &81.9 &82.4 &79.0 &87.7 &77.3 &90.8 &71.8 &91.0 &85.9 &83.7 &95.3 &71.6 &94.1 &81.3 &58.7 &76.4 &82.6 \\
		&SyncSpecCNN\cite{yi2017syncspeccnn} &84.7 &82.0 &81.6 &81.7 &81.9 &75.2 &90.2 &74.9 &\textbf{93.0} &86.1 &84.7 &95.6 &66.7 &92.7 &81.6 &60.6 &82.9 &82.1 \\
		&Pd-Network\cite{klokov2017escape} &85.5 &82.7 &83.3 &82.4 &87.0 &77.9 &90.9 &76.3 &91.3 &87.3 &84.0 &95.4 &68.7 &94.0 &82.9 &63.0 &76.4 &83.2 \\
		\hline
		\multirow{7}*{2018} &KCNet\cite{shen2018mining} &83.7 &82.2 &82.8 &81.5 &86.4 &77.6 &90.3 &76.8 &91.0 &87.2 &84.5 &95.5 &69.2 &94.4 &81.6 &60.1 &75.2 &81.3 \\
		&SO-Net\cite{li2018so} &84.9 &81.0 &82.8 &77.8 &88.0 &77.3 &90.6 &73.5 &90.7 &83.9 &82.8 &94.8 &69.1 &94.2 &80.9 &53.1 &72.9 &83.0 \\
		&RSNet\cite{huang2018recurrent} &84.9 &81.4 &82.7 &86.4 &84.1 &78.2 &90.4 &69.3 &91.4 &87.0 &83.5 &95.4 &66.0 &92.6 &81.8 &56.1 &75.8 &82.2 \\
		&PCNN\cite{atzmon2018point} &85.1 &81.8 &82.4 &80.1 &85.5 &79.5 &90.8 &73.2 &91.3 &86.0 &85.0 &95.7 &73.2 &94.8 &83.3 &51.0 &75.0 &81.8 \\
		&SpiderCNN\cite{xu2018spidercnn} &85.3 &81.7 &83.5 &81.0 &87.2 &77.5 &90.7 &76.8 &91.1 &87.3 &83.3 &95.8 &70.2 &93.5 &82.7 &59.7 &75.8 &82.8 \\
		&SGPN\cite{wang2018sgpn} &85.8 &82.8 &80.4 &78.6 &78.8 &71.5 &88.6 &78.0 &90.9 &83.0 &78.8 &95.8 &77.8 &93.8 &87.4 &60.1 &\textbf{92.3} &\textbf{89.4} \\
		&PointCNN\cite{li2018pointcnn} &86.1 &84.6 &84.1 &\textbf{86.5} &86.0 &80.8 &90.6 &79.7 &92.3 &88.4 &85.3 &96.1 &77.2 &95.3 &84.2 &64.2 &80.0 &83.0 \\
		\hline
		\multirow{4}*{2019} &DGCNN\cite{wang2019dynamic} &84.9 &81.4 &82.7 &86.4 &84.1 &78.2 &90.4 &69.3 &91.4 &87.0 &83.5 &95.4 &66.0 &92.6 &81.8 &56.1 &75.8 &82.2 \\
		&RS-CNN\cite{liu2019relation} &86.2 &84.0 &83.5 &84.8 &88.8 &79.6 &91.2 &\textbf{81.1} &91.6 &88.4 &\textbf{86.0} &96.0 &73.7 &94.1 &83.4 &60.5 &77.7 &83.6 \\
		&KPConv rigid\cite{thomas2019kpconv} &86.2 &85.0 &83.8 &86.1 &88.2 &\textbf{81.6} &91.0 &80.1 &92.1 &87.8 &82.2 &\textbf{96.2} &77.9 &95.7 &\textbf{86.8} &65.3 &81.7 &83.6 \\
		& KPConv deform\cite{thomas2019kpconv} &\textbf{86.4} &\textbf{85.1} &\textbf{84.6} &86.3 &87.2 &81.1 &91.1 &77.8 &92.6 &88.4 &82.7 &\textbf{96.2} &\textbf{78.1} &\textbf{95.8} &85.4 &\textbf{69.0} &82.0 &83.6 \\
		\hline
		\multirow{1}*{2020} &3D-GCN\cite{lin2020convolution} &85.1 &82.1 &83.1 &84.0 &86.6 &77.5 &90.3 &74.1 &90.0 &86.4 &83.8 &95.6 &66.8 &94.8 &81.3 &59.6 &75.7 &82.8 \\
		\hline
		\multirow{1}*{2021} &PAConv\cite{xu2021paconv} &86.1 &84.6 &84.3 &85.0 &\textbf{90.4} &79.7 &90.6 &80.8 &92.0 &\textbf{88.7} &82.2 &95.9 &73.9 &94.7 &84.7 &65.9 &81.4 &84.0 \\
		\hline
		&CRFConv (ours) &85.5 &83.5 &83.9 &84.8 &83.0 &80.2 &\textbf{91.8} &77.9 &91.8 &86.9 &84.9 &95.6 &77.8 &95.6 &82.0 &64.4 &75.3 &80.8\\
		\bottomrule
	\end{tabular}
	}
	\label{tab2}
\end{table}

\subsection{Scene-Level Evaluation}
\subsubsection{Data Preparation}

\paragraph{Conventional Implementation}
Scene-level point cloud data are more complex than object-level data. The former is spatially continuous and usually nonuniform in density and cannot be processed by the network directly. Following \cite{qi2017pointnet,qi2017pointnet++,li2018pointcnn}, we sample blocks from a scene point cloud in training and testing. For indoor scenes, the point cloud is first collected by rooms. Then, each block is randomly sampled from each room. For outdoor scenes, we directly sample blocks from each scene. To control the block size, we then uniformly sample $N$ points in each block to feed the network. In training, the blocks are randomly sampled from the training set in each epoch. In the test, we sample multiple blocks in the test set until each point is evaluated at least once. After that, the final per-point labels can be obtained by voting. In this strategy, neighbor searching and cloud sampling are computed directly in the network. We adopt this strategy as a conventional implementation in scene-level evaluation experiments.
\paragraph{Advanced Implementation}
In conventional implementations, the network will require considerable computation to organized the structure of the point cloud (i.e., neighbor searching and cloud sampling). Therefore, some methods\cite{thomas2019kpconv,hu2020randla} adopt a more computationally efficient method to organize the input point cloud blocks. In this strategy, the multiscale point cloud and the point neighbors at each scale are efficiently precomputed on the CPU. This strategy allows them to handle ten times the input size than the conventional implementation under the same time complexity, which is helpful for training the network on large-scale point clouds. To compare with them, we also employ this strategy as an advanced implementation for the proposed method and conduct experiments on scene-level benchmarks. The advanced implementation is denoted as \textit{CRFConv+} to be distinct from the conventional implementation, which is denoted as \textit{CRFConv}.

\subsubsection{Indoor Scene Segmentation}
The proposed method is evaluated on the S3DIS dataset for scene-level segmentation tasks. We adopt the normalized coordinates in the block and the corresponding RGB color as the input feature. We set the input point size to 8,192 for the conventional implementation (CRFConv) and 40,960 for the advanced implementation (CRFConv+). We take the overall accuracy (OA), mean accuracy (mACC), mean IoU (mIoU), and per class IoU as metrics to evaluate the performance. Since previous methods usually give the results on Area 5, we also report the results on this area in Table \ref{tab3}. From the experimental results, our method can achieve state-of-the-art performance with the precomputed multiscale strategy. For conventional implementation, the proposed method can also outperform some previous methods that prepare data in a conventional style, such as PointNet and PointCNN.
\begin{table}[!h]
	\caption{Performance of the proposed method on the Stanford S3DIS dataset (Area 5).}
	\footnotesize
	\renewcommand\arraystretch{1.0}
	\setlength{\tabcolsep}{0.5mm}
	\centering
	\resizebox{\textwidth}{!}
	{
	\begin{tabular}{c|ccccccccccccccccc}
		\toprule
		&Methods &\textbf{OA} &\textbf{mACC} &\textbf{mIoU} &ceil. &floor &wall &beam &col. &wind. &door &table &chair &sofa &book. &board &clut.\\
		\midrule
		\multirow{2}*{2017} &PointNet\cite{qi2017pointnet} &- &49.0 &41.1 &88.8 &97.3 &69.8 &0.1 &3.9 &46.3 &10.8 &58.9 & 52.6 &5.9 &40.3 &26.4 &32.2 \\
		&SegCloud\cite{tchapmi2017segcloud} &- &57.4 &48.9 &90.1 &96.1 &69.9 &0.0 &18.4 &38.4 &23.1 &70.4 &75.9 &40.9 &58.4 &13.0 &41.6 \\
		\hline
		\multirow{6}*{2018} &Eff 3D Conv\cite{zhang2018efficient} &- &68.3 &51.8 &79.8 &93.9 &69.0 &0.2 &28.3 &38.5 &48.3 &73.6 &71.1 &59.2 &48.7 &29.3 &33.1 \\
		&TangentConv\cite{tatarchenko2018tangent} &- &62.2 &52.6 &90.5 &97.7 &74.0 &0.0 &20.7 &39.0 &31.3 &77.5 &69.4 &57.3 &38.5 &48.8 &39.8 \\
		&RNN Fusion\cite{Xiaoqing20183D} &- &63.9 &57.3 &92.3 &98.2 &79.4 &0.0 &17.6 &22.8 &62.1 &80.6 &74.4 &66.7 &31.7 &62.1 &56.7 \\
		&PointCNN\cite{li2018pointcnn} &85.9 &63.9 &57.3 &92.3 &98.2 &79.4 &0.0 &17.6 &22.8 &62.1 &74.4 &80.6 &31.7 &66.7 &62.1 &56.7 \\
		&SPGGraph\cite{wang2018sgpn} &86.4 &66.5 &58.0 &89.4 &96.9 &78.1 &0.0 &\textbf{42.8} &48.9 &61.6 &\textbf{84.7} &75.4 &69.8 &52.6 &2.1 &52.2 \\
		&PCNN\cite{atzmon2018point} &- &67.0 &58.3 &92.3 &96.2 &75.9 &\textbf{0.3} &6.0 &\textbf{69.5} &63.5 &66.9 &65.6 &47.3 &68.9 &59.1 &46.2 \\
		\hline
		\multirow{4}*{2019} &GACNet\cite{wang2018deep} &87.8 &- &62.9 &92.3 &98.3 &81.9 &0.0 &20.4 &59.1 &40.9 &78.5 &85.8 &61.7 &70.8 &\textbf{74.7} &52.8 \\
		&PointWeb\cite{zhao2019pointweb} &87.0 &66.6 &60.3 &92.0 &98.5 &79.4 &0.0 &21.1 &59.7 &34.8 &76.3 &88.3 &46.9 &69.3 &64.9 &52.5 \\
		&KPConv rigid\cite{thomas2019kpconv} &- &70.9 &65.4 &92.6 &97.3 &81.4 &0.0 &16.5 &54.5 &69.5 &80.2 &90.1 &66.4 &74.6 &63.7 &58.1 \\
		&KPConv deform\cite{thomas2019kpconv} &- &72.8 &\textbf{67.1} &92.8 &97.3 &82.4 &0.0 &23.9 &58.0 &69.0 &81.5 &\textbf{91.0} &\textbf{75.4} &\textbf{75.3} &66.7 &\textbf{58.9} \\
		\hline 
		\multirow{3}*{2020} &FPConv\cite{lin2020fpconv} &- &- &62.8 &\textbf{94.6} &98.5 &80.9 &0.0 &19.1 &60.1 &48.9 &80.6 &88.0 &53.2 &68.4 &68.2 &54.9\\
		&SegGCN &88.2 &70.4 &63.6 &93.7 &\textbf{98.6} &80.6 &0.0 &28.5 &42.6 &\textbf{74.5} &80.9 &88.7 &69.0 &71.3 &44.4 &54.3 \\
		&PointASNL\cite{yan2020pointasnl} &87.7 &68.5 &62.6 &94.3 &98.4 &79.1 &0.0 &26.7 &55.2 &66.2 &83.3 &86.8 &47.6 &68.3 &56.4 &52.1 \\
		\hline
		\multirow{2}*{2021} &PAConv\cite{xu2021paconv} &- &72.3 &65.6 &93.1 &98.4 &\textbf{82.6} &0.0 &22.6 &61.3 &63.3 &78.5 &88.0 &64.5 &73.5 &70.1 &57.3 \\
		&BCM+AFM\cite{qiu2021semantic} &88.9 &73.1 &65.4 &92.9 &97.9 &82.3 &0.0 &23.1 &65.5 &64.9 &78.5 &87.5 &61.4 &70.7 &68.7 &57.2\\
		\hline
		&CRFConv (ours) &88.4 &68.6 &62.0 &94.0 &97.5 &81.8 &0.0 &21.7 &62.2 &45.4 &78.1 &85.2 &51.1 &71.8 &61.7 &55.7 \\
		&CRFConv+ (ours) &\textbf{89.2} &\textbf{73.7} &66.2 &93.3 &96.3 &82.2 &0.0 &23.7 &60.3 &68.2 &82.4 &86.0 &63.4 &73.8 &72.4 &\textbf{58.9} \\
		\bottomrule
	\end{tabular}
	}
	\label{tab3}
\end{table}

\subsubsection{Outdoor Scene Segmentation}
We further evaluate the proposed method on the large-scale outdoor 3D point cloud segmentation benchmark Semantic3D. We first uniformly downsample the training set before sampling blocks to maintain a voxel resolution of $0.01m^3$, which is the as the reduced test set. The normalized position in the block and the corresponding RGB color are taken as the input features for the network. The input point cloud sizes are set to 8,192 and 65,536 for the conventional and advanced implementations, respectively. Because Semantic3D is an online benchmark, we submit the experimental results on the reduced-8 test set to the benchmark website.\footnote{\url{http://semantic3d.net/view_results.php?chl=2}} The experimental results are also shown in Table \ref{tab5}. Note that only the published methods are listed in the table. From the table, the performance of RandLA-Net and SCF-Net is state-of-the-art overall. Next, the proposed method performs comparably with KPConv and BCM\_AFM, and outperforms the other methods.

\begin{table}[!h]
	\caption{Performance of the proposed method on the Semantic3D dataset (reduced-8).}
	\footnotesize
	\renewcommand\arraystretch{1.0}
	\setlength{\tabcolsep}{0.5mm}
	\centering
	\resizebox{\textwidth}{!}
	{
	\begin{tabular}{c|cccccccccccc}
		\toprule
		&Methods &\textbf{OA} &\textbf{mACC} &\textbf{mIoU} &\makecell[c]{man-made\\terrain} &\makecell[c]{natural\\terrain} &\makecell[c]{high\\vegetation} &\makecell[c]{low\\vegetation} &buildings &\makecell[c]{hard\\scape} &\makecell[c]{scanning\\artefacts} &cars \\
		\midrule
		\multirow{2}*{2017} &SnapNet\cite{boulch2017unstructured} &88.6 &70.8 &59.1 &82.0 &77.3 &79.7 &22.9 &91.1 &18.4 &37.3 &64.4 \\
		&SEGCloud\cite{tchapmi2017segcloud} &88.1 &73.1 &61.3 &83.9 &66.0 &86.0 &40.5 &91.1 &30.9 &27.5 &64.3 \\
		\hline
		\multirow{3}*{2018} &RF\_MSSF\cite{thomas2018semantic} &90.3 &71.6 &62.7 &87.6 &80.3 &81.8 &36.4 &92.2 &24.1 &42.6 &56.6 \\
		&MSDVN\cite{roynard2018classification} &88.4 &77.2 &65.3 &83.0 &67.2 &83.8 &36.7 &92.4 &31.3 &50.0 &78.2 \\
		&SPGraph\cite{wang2018sgpn} &94.0 &79.6 &73.2 &97.4 &92.6 &87.9 &44.0 &93.2 &31.0 &63.5 &76.2 \\
		\hline
		\multirow{3}*{2019} &ShellNet\cite{zhang2019shellnet} &93.2 &79.7 &69.3 &96.3 &90.4 &83.9 &41.0 &94.2 &34.7 &43.9 &70.2 \\
		&GACNet\cite{Lei2019Graph} &91.9 &79.5 &70.8 &86.4 &77.7 &\textbf{88.5} &\textbf{60.6} &94.2 &37.3 &43.5 &77.8 \\
		&KPConv\cite{thomas2019kpconv} &92.9 &81.4 &74.6 &90.9 &82.2 &84.2 &47.9 &94.9 &40.0 &\textbf{77.3} &79.7 \\
		\hline
		\multirow{2}*{2020} &RandLA-Net\cite{hu2020randla} &\textbf{94.8} &84.1 &77.4 &95.6 &91.4 &86.6 &51.5 &\textbf{95.7} &\textbf{51.5} &69.8 &76.8 \\
		&FGCN\cite{ma2020global} &- &\textbf{89.3} &62.4 &90.3 &65.2 &86.2 &38.7 &90.1 &31.6 &28.8 &68.2 \\
		\hline
		\multirow{2}*{2021} &BCM+AFM\cite{qiu2021semantic} &94.3 &- &75.3 &96.3 &\textbf{93.7} &87.7 &48.1 &94.6 &43.8 &58.2 &79.5 \\
		&SCF-Net\cite{fan2021scf} &94.7 &85.2 &\textbf{77.6} &97.1 &91.8 &86.3 &51.2 &95.3 &50.5 &67.9 &80.7 \\
		\hline
		&CRFConv (ours) &93.5 &78.4 &72.5 &95.0 &93.1 &86.4 &40.8 &93.3 &33.7 &60.5 &76.8 \\
		&CRFConv+ (ours) &94.2 &84.1 &74.9 &\textbf{98.1} &91.1 &81.8 &44.8 &95.2 &40.8 &59.4 &\textbf{88.0} \\
		\bottomrule
	\end{tabular}
	}
	\label{tab5}
\end{table}

\subsubsection{Discussion}
According to the experimental results for scene-level evaluation, the proposed method can perform well in both indoor and outdoor scenes, which demonstrates the robustness of our method. Overall, the proposed method performs better in indoor scenes because the outdoor scenes are usually more complex and biased than indoor scenes. In Table \ref{tab5}, we find that our method performs better on large or regular objects such as terrain and cars but worse on small or irregular objects such as vegetation and landscape. The reason can be explained by the fact that the former will benefit from the proposed CRF model, but the latter may be erroneously viewed as noise and smoothed by the CRF model in feature extraction, especially when the training samples involving the case are very limited. Actually, this phenomenon occurs not only in this dataset but will be magnified in a biased situation. We will try to address this problem in future works. In addition, we find that the advanced implementations with a precomputed multiscale strategy can largely improve the performance on both indoor and outdoor scenes compared with the conventional implementations. This is because the block sampling strategy damages the completeness of the continuous point cloud, especially when the block size is small, which may harm the overall performance in some way. In contrast, the precomputed multiscale strategy allow us to handle a larger input block size, which is helpful for network learning on scene-level point clouds. More visualization results on the scene-level datasets can be found in the supplementary material.

\subsection{Ablation Study}
To further evaluate the performance of the continuous/discrete CRF convolution, we conduct an ablation study on the S3DIS dataset for the proposed method. First, we compare the performance of the proposed method with/without the continuous/discrete CRF convolution. Table \ref{tab7} shows the experimental results on Area 5. Note that the results in this section are conventionally implemented without merging labels. We adopt the same encoder for all models. In the baseline model, we remove the continuous CRF convolution in the decoder and only perform linear interpolation for point cloud upsampling. Baseline+discrete CRF denotes that we add the discrete CRF at the end of the baseline model. Baseline+continuous CRF employs continuous CRF convolution in the decoder, which is the general version of the proposed method. For baseline+dual CRF, we add the discrete CRF convolution at the end of the general version to form the dual CRF version of the proposed method.

\begin{table}[!h]
	\caption{Performance of the proposed method under different module configurations.}
	\footnotesize
	\renewcommand\arraystretch{1.0} 
	\setlength{\tabcolsep}{2.5mm}
	\centering
	\resizebox{0.7\textwidth}{!}
	{
	\begin{tabular}{cccc}
		\toprule
		Models &\textbf{OA} &\textbf{mACC} &\textbf{mIoU} \\
		\midrule
		Baseline &89.78 &66.25 &59.36 \\
		Baseline+Discrete CRF &90.01 &66.51 &59.88 \\
		Baseline+Continuous CRF (general version) &90.13 &67.91 &61.22 \\
		Baseline+Dual CRF (dual CRF version) &\textbf{90.31} &\textbf{68.12} &\textbf{61.54} \\
		\bottomrule
	\end{tabular}
	}
	\label{tab7}
\end{table}

\begin{figure}[!h]
	\centerline{\includegraphics[width=\columnwidth]{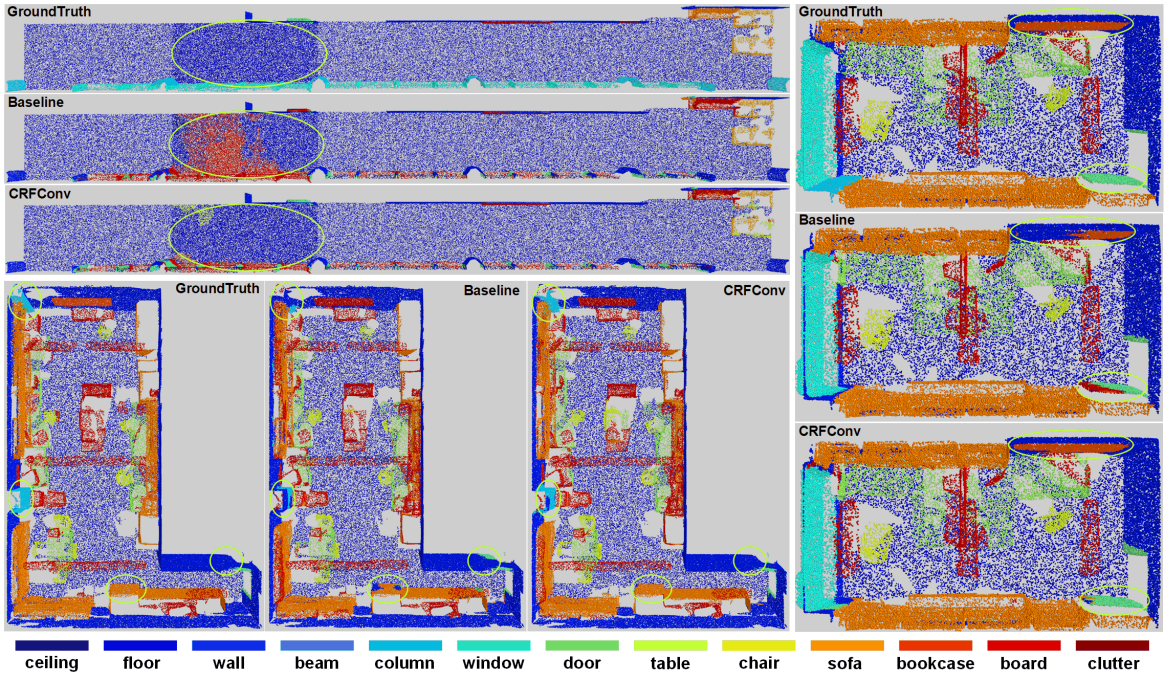}}
	\caption{Visualization comparisons of the baseline model and the CRFConv model. We show three rooms in area 5 of the S3DIS dataset, that is, \textit{hallway\_1} on the top left, \textit{office\_2} on the right, and \textit{office\_40} on the bottom left. The results are visualized in the bird view from the upper axis. Note that the "ceilings" category is removed manually for better visualization. The regions marked by green circles show some significant local differences.} 
	\label{fig11}
\end{figure}

In Table \ref{tab7}, even the baseline model achieves promising performance, which can also prove the effectiveness of the encoder we designed. Compared with the baseline model, the model with continuous CRF has achieves a significant improvement. Although the discrete CRF also improves the performance in some way, the improvement effect is marginal. In traditional discrete CRF, the performance of the model depends largely on the unary term. In other words, the misclassified regions caused by the classifier are hard to correct by the discrete CRF. However, our continuous CRF model can fundamentally improve the presentation (location) ability of the point features, therefore avoiding such discrete CRF model limitations. Additionally, some visualization comparisons with/without the CRFConv are illustrated in Figure \ref{fig11}. We can see that the segmentation results of the CRFConv model look smoother and more complete than those of the baseline model in most cases, especially on some boundary areas (see \textit{office\_2} and \textit{office\_ 40}). Some misclassified regions produced by the baseline model can also be corrected by the proposed CRFConv (see \textit{hallway\_ 1}).

\begin{table}[!h]
	\caption{Performance of RandLA-Net with/without CRFConv on S3DIS (Area 5).}
	\footnotesize
	\renewcommand\arraystretch{1.0}
	\setlength{\tabcolsep}{0.5mm}
	\centering
	\resizebox{0.8\textwidth}{!}
	{
		\begin{tabular}{ccccccccccccccccc}
			\toprule
			Methods &\textbf{mIoU} &ceil. &floor &wall &beam &col. &wind. &door &table &chair &sofa &book. &board &clut.\\
			\midrule
			RandLA-Net\cite{hu2020randla} &63.0 &91.5 &97.0 &81.1 &\textbf{0.0} &22.5 &60.9 &44.8 &\textbf{78.3} &87.7 &\textbf{69.1} &71.1 &65.2 & 50.0 \\
			RandLA-Net+CRFConv &\textbf{64.1} &\textbf{92.5} &\textbf{97.8} &\textbf{82.1} &\textbf{0.0} &\textbf{23.2} &\textbf{62.0} &\textbf{49.9} &78.1 &\textbf{87.8} &56.9 &\textbf{72.1} &\textbf{77.3} &\textbf{53.6} \\
			\bottomrule
		\end{tabular}
	}
	\label{tab8}
\end{table}
The proposed CRFConv layer is flexible and scalable and can be used as a plug-and-play module for integrating into existing encoder-decoder networks to further improve their performance. In this section, we integrate CRFConv with RandLA-Net\cite{hu2020randla}, which is one of the representative methods in the community, and conduct experiments on the S3DIS dataset. Specifically, we package the CRFConv as an upsampling layer and insert it into the decoding layers of the original RandLA-Net. All other settings follow the same schemes as the original paper. We report the results on Area5 in Table \ref{tab8}. In the table, our CRFConv can further improve its performance, which also proves the effectiveness of the proposed CRFConv. Through this experiment, we demonstrate that the proposed method has a certain generality that can improve the performance of the network independent of specific encoders.

\subsection{Parameter Analysis}
\begin{figure}[!h]
	\centerline{\includegraphics[width=0.5\columnwidth]{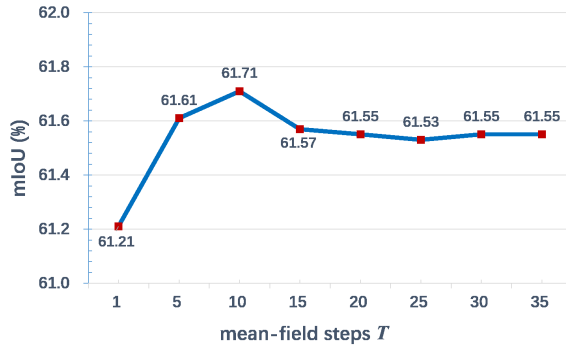}}
	\caption{The test performance of the CRFConv on various mean-field steps.} 
	\label{fig7}
\end{figure}

The mean-field step $T$ of our CRFConv is a flexible parameter, which allows us to adopt different mean-field steps for the training and test phases. In training, we set the mean-field step as 1 for efficiency and to avoid vanishing/exploding gradients. Therefore, we evaluate the performance of various mean-field steps in the test phase. The performance of different mean-field steps on the validation set is illustrated in Figure \ref{fig7}. According to this figure, the continuous CRF performs best when the mean-field step is approximately 10. With the increase in the mean-field steps, the performance tends gradually become saturated. Owing to the fidelity ability of CRFConv, we do not observe obvious performance degradation caused by oversmoothness, which may occur in most current popular graph convolutions. However, we find that the model performs best when it is about to converge rather than converge. When the model converges, the feature may become smoother, which may be harmful to some small or irregular objects. This can explain why the performance decreases slightly when the CRF reaches convergence. This experimental result is also coincident with our analysis of the outdoor scenes.

\subsection{Time complexity}
In this section, we conduct a runtime analysis of the proposed method. First, the runtime of the proposed method with/without the precomputed multiscale strategy is given. Then, we additionally make a runtime comparison with some previous methods. Specifically, we choose the S3DIS benchmark for this experiment and run the previous methods on our experimental platform. The experimental setting follows the original paper. We give both the training and test runtime of each iteration including data preparation. To make a fair comparison, we make the input point cloud size of our method the same as previous methods. Note that, because KPConv adopts various input sizes, we report an approximate value here. The experimental results are shown in Table \ref{tab10}. From the table, we find that PointNet is the most efficient method because of its simple architecture. Although the overall runtime of the proposed method is higher than that of previous methods, the proposed CRFConv module only adds a small runtime cost to the backbone network in both training and testing. Moreover, due to the precomputed multiscale strategy, the efficiency of the network greatly improved. This strategy allows us to handle a larger batch size but with lower time complexity.
\begin{table}[!h]
	\caption{Runtime analysis of the proposed method.}
	\footnotesize
	\renewcommand\arraystretch{1.0} 
	\setlength{\tabcolsep}{2.0mm}
	\centering
	\resizebox{0.8\textwidth}{!}
	{
	\begin{tabular}{ccccc}
		\toprule
		Methods &Batch Size &\makecell[c]{Precomputed\\ Multiscale} &\makecell[c]{Training Time\\ (ms/iter)} &\makecell[c]{Test Time\\ (ms/iter)} \\
		\hline
		PointNet\cite{qi2017pointnet} &$16\times4096$ &No &138 &98 \\
		PointNet++(SSG)\cite{qi2017pointnet++} &$16\times4096$ &No &532 &495 \\
		PointNet++(MSG)\cite{qi2017pointnet++} &$16\times4096$ &No &694 &633 \\
		KPConv\cite{thomas2019kpconv} &$\approx8\times15000$ &Yes &552 &271 \\
		RandLA-Net\cite{hu2020randla} &$8\times40960$ &Yes &760 &448 \\
		\hline
		\multirow{2}*{Baseline (ours)}		&$8\times8192$ &No &935 &475 \\
		&$8\times40960$ &Yes &909 &625 \\	
		\hline			 		
		\multirow{2}*{Baseline+CRFConv (ours)}	&$8\times8192$ &No &1141 &653 \\
		&$8\times40960$ &Yes &990 &667 \\
		\hline
		RandLA-Net+CRFConv (ours)	&$8\times40960$ &Yes &815 &485 \\
		\bottomrule
	\end{tabular}
	}
	\label{tab10}
\end{table}

\section{Conclusion}
In this paper, we addressed the problem of point cloud segmentation. First, we proposed a continuous CRF graph convolution to capture the structure of the point features, by which the representation ability of the features improved. Then, we designed an encoder-decoder point cloud segmentation network based on the proposed continuous CRF convolution. The continuous CRF convolution embedded in the decoder can restore the details of the high-level features produced by the encoder to enhance the localization ability of the network. Furthermore, we also showed how to formulate the discrete CRF as another message-passing graph convolution and combined it with the continuous CRF convolution, by which we modeled the data affinity not only in feature space but also in label space to further improve the segmentation results. Experiments on various datasets demonstrated the effectiveness of the proposed method.

However, compared with indoor scenes, we find that the proposed method performs worse in outdoor scenes that are usually more complex and biased than the former. In this case, the CRF model erroneously viewed some small or irregular objects as noise in feature extraction, especially when the number of training samples involving the case was very limited. This situation may be relieved by increasing the input point cloud size with the precomputed multiscale strategy. In the future, we will attempt to address this problem. In this work, we focused the decoding of the point cloud features and achieve promising results. However, the overall performance of the network depends on more than one aspect. For example, the feature encoder may play a decisive role. In future work, we will also attempt to investigate more feature encoding methods, such as convolution in the point cloud domain. Moreover, we will continue exploring the feasibility of integrating graph neural networks with probabilistic graphical models.

\section*{Acknowledgment}
This work is partially supported by National Natural Science Foundation of China under Grant Nos 61872188, 61703209, U1713208, 61972204, 61672287, 61861136011, 61773215, and by the French Labex MS2T ANR-11-IDEX-0004-02 through the program Investments for the future.

\bibliography{mybibfile}

\end{document}


\begin{frontmatter}
		
\title{Supplementary Material of Continuous Conditional Random Field Convolution for Point Cloud Segmentation}
		
\author[1,2]{Fei Yang}
\ead{yangfei92516@163.com}
\author[3]{Franck Davoine}
\ead{franck.davoine@hds.utc.fr}			
\author[2]{Huan Wang}
\ead{wanghuanphd@njust.edu.cn}
\author[1,2]{Zhong Jin\corref{cor}}
\ead{zhongjin@njust.edu.cn}

\cortext[cor]{Corresponding author}
\address[1]{Key Laboratory of Intelligent Perception and Systems for High-Dimensional Information of Ministry of Education, \\Nanjing University of Science and Technology, Nanjing 210094, China}
\address[2]{School of Computer Science and Engineering, Nanjing University of Science and Technology, Nanjing 210094, China}
\address[3]{Alliance Sorbonne Universit\'{e}, Universit\'{e} de technologie de Compi\`{e}gne, CNRS, Heudiasyc Lab., CS 60319 - F-60203 Compi\`{e}gne Cedex, France}

\begin{abstract}	
This document provides some supplementary contents for the paper: Continuous Conditional Random Field Convolution for Point Cloud Segmentation. Section 1 of this document presents some details about the proof of Theorem 1. The model-level analysis of the relationship between the proposed CRFConv and previous diffusion-based methods is given in section 2. More experimental details and results can be found in section 3.
\end{abstract}
\end{frontmatter}

\section{The proof of Theorem 1}
\noindent\textit{Problem statement}:\\

Given a continuous CRF defined on graph $\mathcal{G}(\mathcal{V},\mathcal{E})$, its corresponding Gibbs distribution can be written as:
\begin{equation}
P(X)=\frac{1}{Z}\exp(-E(X)),
\label{eq1}
\end{equation}
where the $X=\{\mathbf{x}_i\in\mathbb{R}^d|i\in\mathcal{V}\}$ denotes the set of all latent features, $Z$ is a partition function to ensure a distribution. The energy function $E(X)$ is defined as a continuous quadratic energy model:
\begin{equation}
E(X)=\sum_{i\in\mathcal{V}}(\mathbf{x}_i-\mathbf{z}_i)^\top(\mathbf{x}_i-\mathbf{z}_i)+\sum_{i\in\mathcal{V}}\sum_{j\in\mathcal{N}(i)}(\mathbf{x}_i-\mathbf{x}_j)^\top\mathbf{w}_{ij}(\mathbf{x}_i-\mathbf{x}_j),
\label{eq2}
\end{equation}
where $\mathbf{z}_i$ denotes the observed feature of node $i$. $\mathcal{N}(i)$ denotes the neighbor nodes set of node $i$ in $\mathcal{G}$. $\mathbf{w}_{ij}\in\mathbb{R}^{d\times d}\succ\mathbf{0}$ is the weight matrix of two connected nodes in $\mathcal{G}$.

Let $Q(X)=\prod_iQ_i(\mathbf{x}_i)$ be the approximate distribution of $P(\mathbf{x})$, where $Q_i\sim N(\mathbf{\mu}_i,\mathbf{\Sigma}_i)$. Our object is to obtain $\{(\mathbf{\mu}_i^*,\mathbf{\Sigma}_i^*)|i\in\mathcal{V}\}$ by solving the following problem:
\begin{equation}
\{(\mathbf{\mu}_i^*,\mathbf{\Sigma}_i^*)|i\in\mathcal{V}\}=\mathop{argmin}_{\{(\mathbf{\mu}_i,\mathbf{\Sigma}_i)|i\in\mathcal{V}\}}KL(Q||P),
\label{eq3}
\end{equation}
where 
\begin{equation}
KL(Q||P)=\int Q(X)\log\frac{Q(X)}{P(X)}dX
\label{eq4}
\end{equation}
is the KL-divergence between $Q(X)$ and $P(X)$.

\noindent\textit{Solution}:\\
\begin{align}
\{(\mathbf{\mu}_i^*,\mathbf{\Sigma}_i^*)|i\in\mathcal{V}\}
&=\mathop{argmin}_{\{(\mathbf{\mu}_i,\mathbf{\Sigma}_i)|i\in\mathcal{V}\}}KL(Q||P)\notag\\
&=\mathop{argmin}_{\{(\mathbf{\mu}_i,\mathbf{\Sigma}_i)|i\in\mathcal{V}\}}\int Q(X)\log Q(X)dX-\int Q(X)\log P(X)dX\notag\\
&=\mathop{argmin}_{\{(\mathbf{\mu}_i,\mathbf{\Sigma}_i)|i\in\mathcal{V}\}}\sum_{i\in\mathcal{V}}\int Q_i(\mathbf{x}_i)\log Q_i(\mathbf{x}_i)d\mathbf{x}_i+\int Q(X)E(X)dX+\underbrace{\log Z}_{\text{constant}}\notag\\
&=\mathop{argmin}_{\{(\mathbf{\mu}_i,\mathbf{\Sigma}_i)|i\in\mathcal{V}\}}-\sum_{i\in\mathcal{V}}\mathbb{H}_{Q_i}[\mathbf{x}_i]+\mathbb{E}_{X\sim Q}[E(X)]\notag\\
&=\mathop{argmin}_{\{(\mathbf{\mu}_i,\mathbf{\Sigma}_i)|i\in\mathcal{V}\}}-\frac{1}{2}\sum_{i\in\mathcal{V}}\bigg(\log |\mathbf{\Sigma_i}|+\underbrace{d(\log 2\pi+1)}_{\text{constant}}\bigg)+\mathbb{E}_{X\sim Q}[E(X)]\notag\\
&=\mathop{argmin}_{\{(\mathbf{\mu}_i,\mathbf{\Sigma}_i)|i\in\mathcal{V}\}}-\frac{1}{2}\sum_{i\in\mathcal{V}}\log |\mathbf{\Sigma_i}|+\sum_{i\in\mathcal{V}}\mathbb{E}_{x\sim Q}[\mathbf{x}_i^\top\mathbf{x}_i]-2\sum_{i\in\mathcal{V}}\mathbb{E}_{x\sim Q}[\mathbf{z}_i^\top\mathbf{x}_i]+\underbrace{\sum_{i\in\mathcal{V}}\mathbb{E}_{x\sim Q}[\mathbf{z}_i^\top\mathbf{z}_i]}_{\text{constant}}\notag\\
&+\sum_{i\in\mathcal{V}}\sum_{j\in\mathcal{N}(i)}\mathbb{E}_{x\sim Q}[\mathbf{x}_i^\top\mathbf{w}_{ij}\mathbf{x}_i]-2\sum_{i\in\mathcal{V}}\sum_{j\in\mathcal{N}(i)}\mathbb{E}_{x\sim Q}[\mathbf{x}_i^\top\mathbf{w}_{ij}\mathbf{x}_j]+\sum_{i\in\mathcal{V}}\sum_{j\in\mathcal{N}(i)}\mathbb{E}_{x\sim Q}[\mathbf{x}_j^\top\mathbf{w}_{ij}\mathbf{x}_j]\notag\\
&=\mathop{argmin}_{\{(\mathbf{\mu}_i,\mathbf{\Sigma}_i)|i\in\mathcal{V}\}}-\frac{1}{2}\sum_{i\in\mathcal{V}}\log |\mathbf{\Sigma_i}|+\sum_{i\in\mathcal{V}}\mathbb{E}_{x\sim Q}[\text{tr}(\mathbf{x}_i^\top\mathbf{x}_i)]-2\sum_{i\in\mathcal{V}}\mathbb{E}_{x\sim Q}[\mathbf{z}_i^\top\mathbf{x}_i]+\sum_{i\in\mathcal{V}}\sum_{j\in\mathcal{N}(i)}\mathbb{E}_{x\sim Q}[\text{tr}(\mathbf{x}_i^\top\mathbf{w}_{ij}\mathbf{x}_i)]\notag\\
&-2\sum_{i\in\mathcal{V}}\sum_{j\in\mathcal{N}(i)}\mathbb{E}_{x\sim Q}[\text{tr}(\mathbf{x}_i^\top\mathbf{w}_{ij}\mathbf{x}_j)]+\sum_{i\in\mathcal{V}}\sum_{j\in\mathcal{N}(i)}\mathbb{E}_{x\sim Q}[\text{tr}(\mathbf{x}_j^\top\mathbf{w}_{ij}\mathbf{x}_j)]\notag\\
&=\mathop{argmin}_{\{(\mathbf{\mu}_i,\mathbf{\Sigma}_i)|i\in\mathcal{V}\}}-\frac{1}{2}\sum_{i\in\mathcal{V}}\log |\mathbf{\Sigma_i}|+\sum_i\mathbb{E}_{x\sim Q}[\text{tr}(\mathbf{x}_i\mathbf{x}_i^\top)]-2\sum_{i\in\mathcal{V}}\mathbb{E}_{x\sim Q}[\mathbf{z}_i^\top\mathbf{x}_i]+\sum_{i\in\mathcal{V}}\sum_{j\in\mathcal{N}(i)}\mathbb{E}_{x\sim Q}[\text{tr}(\mathbf{x}_i\mathbf{x}_i^\top\mathbf{w}_{ij})]\notag\\
&-2\sum_{i\in\mathcal{V}}\sum_{j\in\mathcal{N}(i)}\mathbb{E}_{x\sim Q}[\text{tr}(\mathbf{x}_j\mathbf{x}_i^\top\mathbf{w}_{ij})]+\sum_{i\in\mathcal{V}}\sum_{j\in\mathcal{N}(i)}\mathbb{E}_{x\sim Q}[\text{tr}(\mathbf{x}_j\mathbf{x}_j^\top\mathbf{w}_{ij})]\notag\\
&=\mathop{argmin}_{\{(\mathbf{\mu}_i,\mathbf{\Sigma}_i)|i\in\mathcal{V}\}}-\frac{1}{2}\sum_{i\in\mathcal{V}}\log |\mathbf{\Sigma_i}|+\sum_{i\in\mathcal{V}}\text{tr}(\mathbb{E}_{x\sim Q}[\mathbf{x}_i\mathbf{x}_i^\top])-2\sum_{i\in\mathcal{V}}\mathbb{E}_{x\sim Q}[\mathbf{z}_i^\top\mathbf{x}_i]+\sum_{i\in\mathcal{V}}\sum_{j\in\mathcal{N}(i)}\text{tr}(\mathbb{E}_{x\sim Q}[\mathbf{x}_i\mathbf{x}_i^\top]\mathbf{w}_{ij})\notag\\
&-2\sum_{i\in\mathcal{V}}\sum_{j\in\mathcal{N}(i)}\text{tr}(\mathbb{E}_{x\sim Q}[\mathbf{x}_j\mathbf{x}_i^\top]\mathbf{w}_{ij})+\sum_{i\in\mathcal{V}}\sum_{j\in\mathcal{N}(i)}\text{tr}(\mathbb{E}_{x\sim Q}[\mathbf{x}_j\mathbf{x}_j^\top]\mathbf{w}_{ij})\notag\\
&=\mathop{argmin}_{\{(\mathbf{\mu}_i,\mathbf{\Sigma}_i)|i\in\mathcal{V}\}}-\frac{1}{2}\sum_{i\in\mathcal{V}}\log |\mathbf{\Sigma_i}|+\sum_{i\in\mathcal{V}}\text{tr}(\mathbf{\Sigma}_i+\mathbf{\mu}_i\mathbf{\mu}_i^\top)-2\sum_{i\in\mathcal{V}}\mathbf{z}_i^\top\mathbf{\mu}_i+\sum_{i\in\mathcal{V}}\sum_{j\in\mathcal{N}(i)}\text{tr}((\mathbf{\Sigma}_i+\mathbf{\mu}_i\mathbf{\mu}_i^\top)\mathbf{w}_{ij})\notag\\
&-2\sum_{i\in\mathcal{V}}\sum_{j\in\mathcal{N}(i)}\text{tr}((\mathbf{\mu}_j\mathbf{\mu}_i^\top)\mathbf{w}_{ij})+\sum_{i\in\mathcal{V}}\sum_{j\in\mathcal{N}(i)}\text{tr}((\mathbf{\Sigma}_j+\mathbf{\mu}_j\mathbf{\mu}_j^\top)\mathbf{w}_{ij})\notag\\
&=\mathop{argmin}_{\{(\mathbf{\mu}_i,\mathbf{\Sigma}_i)|i\in\mathcal{V}\}}\underbrace{\bigg(\sum_{i\in\mathcal{V}}\mathbf{\mu}_i^\top\mathbf{\mu}_i-2\sum_{i\in\mathcal{V}}\mathbf{z}_i^\top\mathbf{\mu}_i+\sum_{i\in\mathcal{V}}\sum_{j\in\mathcal{N}(i)}\mathbf{\mu}_i^\top\mathbf{w}_{ij}\mathbf{\mu}_i-2\sum_{i\in\mathcal{V}}\sum_{j\in\mathcal{N}(i)}\mathbf{\mu}_i^\top\mathbf{w}_{ij}\mathbf{\mu}_j+\sum_{i\in\mathcal{V}}\sum_{j\in\mathcal{N}(i)}\mathbf{\mu}_j^\top\mathbf{w}_{ij}\mathbf{\mu}_j\bigg)}_{\text{temrs consist of $\{\mathbf{\mu}_i|i\in\mathcal{V}\}$}}\notag\\
&+\underbrace{\bigg(-\frac{1}{2}\sum_{i\in\mathcal{V}}\log |\mathbf{\Sigma_i}|+\sum_{i\in\mathcal{V}}\text{tr}(\mathbf{\Sigma}_i)+\sum_{i\in\mathcal{V}}\sum_{j\in\mathcal{N}(i)}\text{tr}(\mathbf{\Sigma}_i\mathbf{w}_{ij})+\sum_{i\in\mathcal{V}}\sum_{j\in\mathcal{N}(i)}\text{tr}(\mathbf{\Sigma}_j\mathbf{w}_{ij})\bigg)}_{\text{temrs consist of $\{\mathbf{\Sigma}_i|i\in\mathcal{V}\}$}}.\notag\\
\label{eq5}
\end{align}

\newpage

To minimize $\mathbf{\mu}_i$ and $\mathbf{\Sigma}_i$, we only need to concern about the terms which contains $\mathbf{\mu}_i$ and $\mathbf{\Sigma}_i$ in Eq.~(\ref{eq5}), respectively. 

For $\mathbf{\mu}_i$, we have
\begin{equation}
\begin{aligned}
\mathbf{\mu}_i^*
&=\mathop{argmin}_{\mathbf{\mu}_i}\mathbf{\mu}_i^\top\mathbf{\mu}_i-2\mathbf{z}_i^\top\mathbf{\mu}_i+\sum_{j\in\mathcal{N}(i)}\mathbf{\mu}_i^\top\mathbf{w}_{ij}\mathbf{\mu}_i-2\sum_{j\in\mathcal{N}(i)}\mathbf{\mu}_i^\top\mathbf{w}_{ij}\mathbf{\mu}_j\\
&=\mathop{argmin}_{\mathbf{\mu}_i}\mathbf{\mu}_i^\top\bigg(\mathbf{I}+\sum_{j\in\mathcal{N}(i)}\mathbf{w}_{ij}\bigg)\mathbf{\mu}_i-2\mathbf{z}_i^\top\mathbf{\mu}_i-2\mathbf{\mu}_i^\top\sum_{j\in\mathcal{N}(i)}\mathbf{w}_{ij}\mathbf{\mu}_j\\
\end{aligned}.
\label{eq6}
\end{equation}

For $\mathbf{\Sigma}_i$, we have
\begin{equation}
\begin{aligned}
\mathbf{\Sigma}_i^*
&=\mathop{argmin}_{\mathbf{\Sigma}_i}-\frac{1}{2}\log |\mathbf{\Sigma_i}|+\text{tr}(\mathbf{\Sigma}_i)+\sum_{j\in\mathcal{N}(i)}\text{tr}(\mathbf{\Sigma}_i\mathbf{w}_{ij})\\
&=\mathop{argmin}_{\mathbf{\Sigma}_i}\text{tr}\bigg(\mathbf{\Sigma}_i\big(\mathbf{I}+\sum_{j\in\mathcal{N}(i)}\mathbf{w}_{ij}\big)\bigg)-\frac{1}{2}\log |\mathbf{\Sigma_i}|
\end{aligned}
\label{eq7}
\end{equation}

Compute the gradients of Eq.~(\ref{eq6}) and Eq.~(\ref{eq7}) with respect to $\mathbf{\mu}_i$ and $\mathbf{\Sigma}_i$ respectively and let the gradients to be zero, we can get the update equations of $\mathbf{\mu}_i^*$ and $\mathbf{\Sigma}_i^*$:
\begin{equation}
\mathbf{\mu}_i^*=\bigg(\mathbf{I}+\sum_{j\in\mathcal{N}(i)}\mathbf{w}_{ij}\bigg)^{-1}\bigg(\mathbf{z}_i+\sum_{j\in\mathcal{N}(i)}\mathbf{w}_{ij}\mathbf{\mu}_j\bigg)
\label{eq8}
\end{equation}
\begin{equation}
\mathbf{\Sigma}_i^*=\frac{1}{2}\bigg(\mathbf{I}+\sum_{j\in\mathcal{N}(i)}\mathbf{w}_{ij}\bigg)^{-1}
\label{eq9}
\end{equation}

To maximize $Q(X)$, we only need to compute $\mathbf{\mu}_i$ for all $Q_i(\mathbf{x}_i)$. It can be found that the update equation of $\mathbf{\mu}_i$ in the mean-field approximation is the same as that of $\mathbf{x}_i$ in the coordinate descent algorithm, by which Theorem 1 can be proven.

\section{The relationship with diffusion}
Consider an anisotropic diffusion process on a heat field $\mathcal{H}$ defined on graph $\mathcal{G}_\mathcal{H}$. Denote $\mathbf{h}_t$ as the heat vector of $\mathcal{H}$ at time $t$ on $\mathcal{G}_\mathcal{H}$. And let $\mathbf{L}_\mathcal{H}=\mathbf{I}-\mathbf{D}_\mathcal{H}^{-1}\mathbf{W}_\mathcal{H}$ be the normalized Laplacian matrix of $\mathcal{G}_\mathcal{H}$, where $\mathbf{D}_\mathcal{H}$ and $\mathbf{W}_\mathcal{H}$ are the degree and adjacent matrix, respectively. The diffusion process on $\mathcal{G}_\mathcal{H}$ can be described as the following PDE with a discrete form:
\begin{equation}
\mathbf{h}^{t+1}-\mathbf{h}^t=-c\mathbf{L}_\mathcal{H}\mathbf{h}^t,
\label{eq18}
\end{equation}
where $c$ is a diffusion coefficient to control the diffusion speed.

In the diffusion process, the heat field is a scalar field. By unfolding Eq.~(\ref{eq18}), we can obtain the temperature $h_i^t$ in the heat field $\mathcal{H}$ for any position $i$ and time $t$:
\begin{equation}
\begin{aligned}
h_i^{t+1}
&=h_i^t-c\sum_{j\in\mathcal{N}(i)}w_{ij}(h_i^t-h_j^t)\\
&=\frac{1}{2}(h_i^t+\sum_{j\in\mathcal{N}(i)}w_{ij}h_j^t) \qquad (\text{by setting}\ c=\frac{1}{2}),
\end{aligned}
\label{eq19}
\end{equation}
where $w_{ij}$ is the weight of two adjacent nodes. If we set the compatibility transform matrix $\mathbf{C}=\mathbf{I}$ in our model, that is, each channel is independent of others, we can find that our CRFConv is equivalent to a diffusion model built on each channel individually at time $t=0$. However, when $t>0$, the difference between the diffusion process and our CRFConv is that the diffusion process adds the previous state at each iteration while our CRFConv always adds the initial state at each iteration.

Furthermore, when the diffusion process reaches the stable state, we have $\mathbf{L}_{\mathcal{H}}\mathbf{h}=\mathbf{0}$,  which is the solution of 
\begin{equation}
\mathbf{h}^*=\mathop{argmin}_{\mathbf{h}}\mathbf{h}^\top\mathbf{L}_{\mathcal{H}}\mathbf{h},
\label{eq20}
\end{equation}
where $\mathbf{h}^{\mathrm{T}}\mathbf{L}_{\mathcal{H}}\mathbf{h}$ is the Dirichlet energy. It can be found that the Dirichlet energy only acts as the smoothness term of our CRF model. That is to say, the diffusion-based methods only ensure the smoothness but ignore the uniqueness of features, which could cause over smoothness, while the proposed method does. In other words, our CRFConv can overcome the over smoothness phenomenon usually occurred in current popular graph convolution methods. Moreover, the diffusion-based methods suppose that each channel is independent of the others, while our CRFConv models the relations between each channel by a compatibility transformation matrix $\mathbf{C}$ which can be learned from the data automatically to improve the representation ability of the model.

\section{More experimental details and results}
\subsection{Implementation details}
The details of the point cloud segmentation network can be found in Table \ref{tab1}. Note that the Batch Normalization with the momentum of 0.98 and the LeakyReLU with the negative slope of 0.1 are added after each convolution layer. For brevity, we did not show them in Table \ref{tab1}. We construct the point cloud graph based on kNN in both the encoder and the decoder. The dilated kNN is only available for the PointConv in the encoder. The momentum stochastic gradient descent (SGD) algorithm is used to minimize the cross entropy-loss, with a momentum of 0.95. We take an exponential decay strategy for the learning rate with initial rate of 0.01. The decay rate is set to ensure the learning rate is divided by 10 every 50 epochs. The network will reach convergence within 100 epochs. The batch sizes are set to 16 and 8 for the conventional and advanced (with precomputed multiscale strategy) implementations, respectively.
\begin{table}[!h]
	\caption{Detailed architecture of the proposed point cloud segmentation network.}
	\renewcommand\arraystretch{1.5}
	\setlength{\tabcolsep}{5.0mm}
	\centering
	\resizebox{0.85\textwidth}{!}
	{
		\begin{tabular}{c|c|c}
			\toprule
			\multicolumn{2}{c|}{\textbf{Modules}}  &\textbf{Layers} \\ 
			\hline 
			\multirow{10}*{\textbf{Encoder}}		&\multirow{2}*{\textbf{conv\_1}} 	&PointConv(in\_dims=F, out\_dims=64,kernel\_size=32, dil=1, ratio=1.0) \\					&&PointConv(in\_dims=64, out\_dims=64, kernel\_size=32, dil=1, ratio=1.0) \\ 
			\cline{2-3}
			&\multirow{2}*{\textbf{conv\_2}} 	&PointConv(in\_dims=64, out\_dims=128, kernel\_size=16, dil=2, ratio=0.25) \\ 					&&PointConv(in\_dims=128, out\_dims=128, kernel\_size=16, dil=2, ratio=1.0) \\ 
			\cline{2-3}
			&\multirow{2}*{\textbf{conv\_3}}	&PointConv(in\_dims=128, out\_dims=256, kernel\_size=16, dil=4, ratio=0.375) \\
												&&PointConv(in\_dims=256, out\_dims=256, kernel\_size=16, dil=4, ratio=1.0) \\
			\cline{2-3}
			&\multirow{2}*{\textbf{conv\_4}}	&PointConv(in\_dims=256, out\_dims=512, kernel\_size=16, dil=4, ratio=0.375) \\
												&&PointConv(in\_dims=512, out\_dims=512, kernel\_size=16, dil=4, ratio=1.0) \\
			\cline{2-3}
			&\multirow{2}*{\textbf{conv\_5}}	&PointConv(in\_dims=512, out\_dims=1024, kernel\_size=16, dil=2, ratio=0.375) \\
												&&PointConv(in\_dims=1024, out\_dims=1024, kernel\_size=16, dil=2, ratio=1.0) \\
			\hline
			\multirow{7}*{\textbf{Decoder}}		
			&\textbf{deconv\_4}					&CRFConv(unary\_dims=1024, pairwise\_dims=512, out=512, kernel\_size=16) \\
			\cline{2-3}
			&\multirow{2}*{\textbf{deconv\_3}}	&Linear(in\_dims=512+512, out\_dims=512) \\
												&&CRFConv(unary\_dims=512, pairwise\_dims=256, out\_dims=256, kernel\_size=16) \\		
			\cline{2-3}
			&\multirow{2}*{\textbf{deconv\_2}}	&Linear(in\_dims=256+256, out\_dims=256) \\
												&&CRFConv(unary\_dims=256, pairwise\_dims=128, out\_dims=128, kernel\_size=16) \\				
			\cline{2-3}
			&\multirow{2}*{\textbf{deconv\_1}}	&Linear(in\_dims=128+128, out\_dims=128) \\	
												&&CRFConv(unary\_dims=128, pairwise\_dims=64, out\_dims=64, kernel\_size=16) \\
			\hline
			\multirow{4}*{\textbf{Classifier}}		&\multirow{2}*{\textbf{fc\_1}}		&Linear(in\_dims=64+64, out\_dims=256) \\ 									&&LeakyReLU(negative\_slope=0.1) \\
			\cline{2-3}
			&\multirow{2}*{\textbf{fc\_2}} 		&Linear(in\_dims=256, out\_dims=L) \\ 									&&Softmax(dim=-1) \\
			\hline
			\multicolumn{2}{c|}{\textbf{Discrete CRF} (optional)}	&DiscreteCRFConv(in\_dims=L, out\_dims=L, kernel\_size=32)\\
			\bottomrule
		\end{tabular}
	}
	\label{tab1}
\end{table}

\clearpage
\subsection{Some visualization results}
\begin{figure}[!h]
	\centerline{\includegraphics[width=\columnwidth]{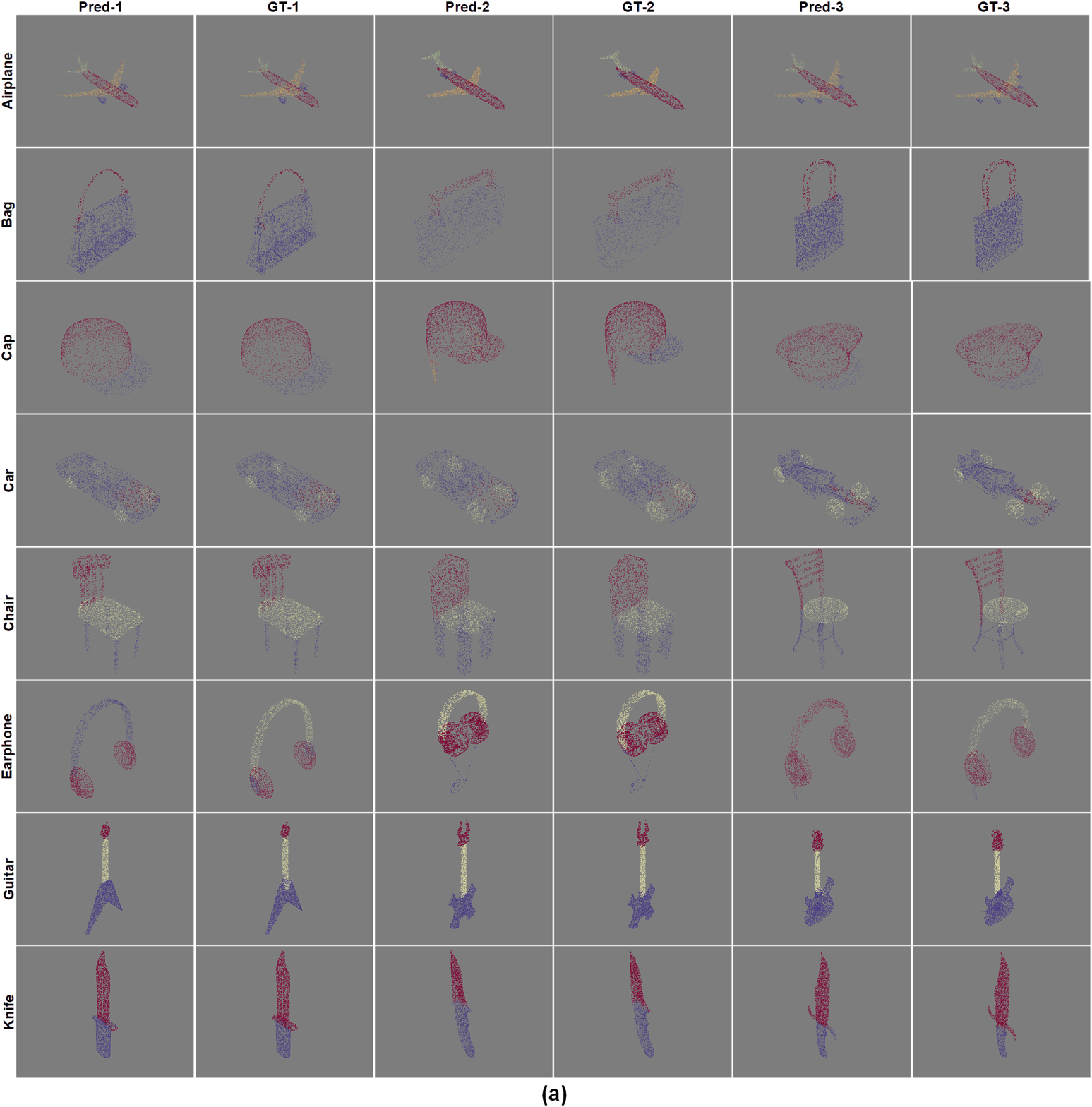}}
\end{figure}
\clearpage
\begin{figure}[!h]
	\centerline{\includegraphics[width=\columnwidth]{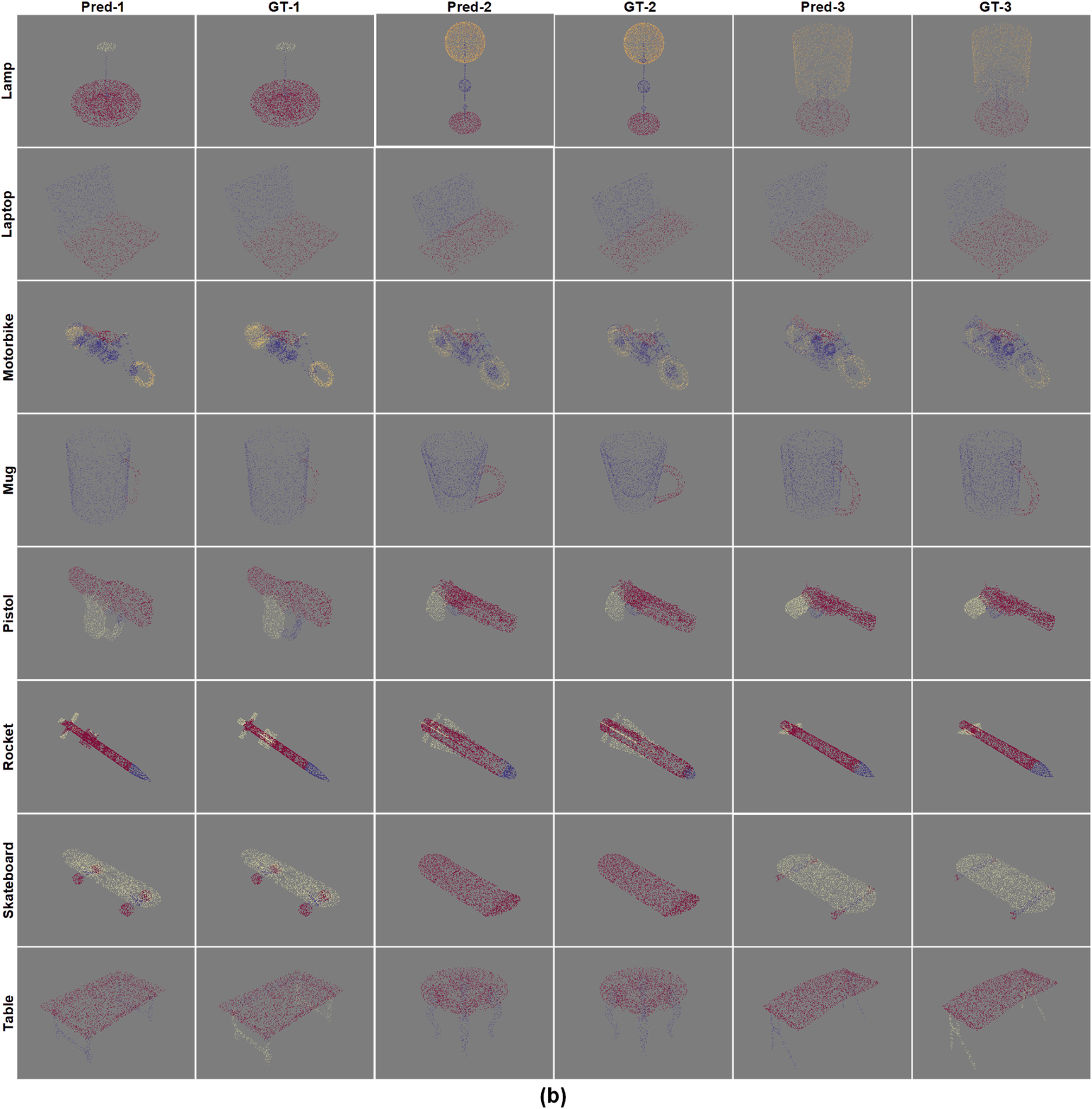}}
	\caption{Visualization results compared with the groundtruth on the ShapeNet Part dataset. Three samples are shown for each object category, in which each part is visualized in a unique color.}
	\label{fig1}
\end{figure}
\clearpage
\begin{figure}[!h]
\centerline{\includegraphics[width=\columnwidth]{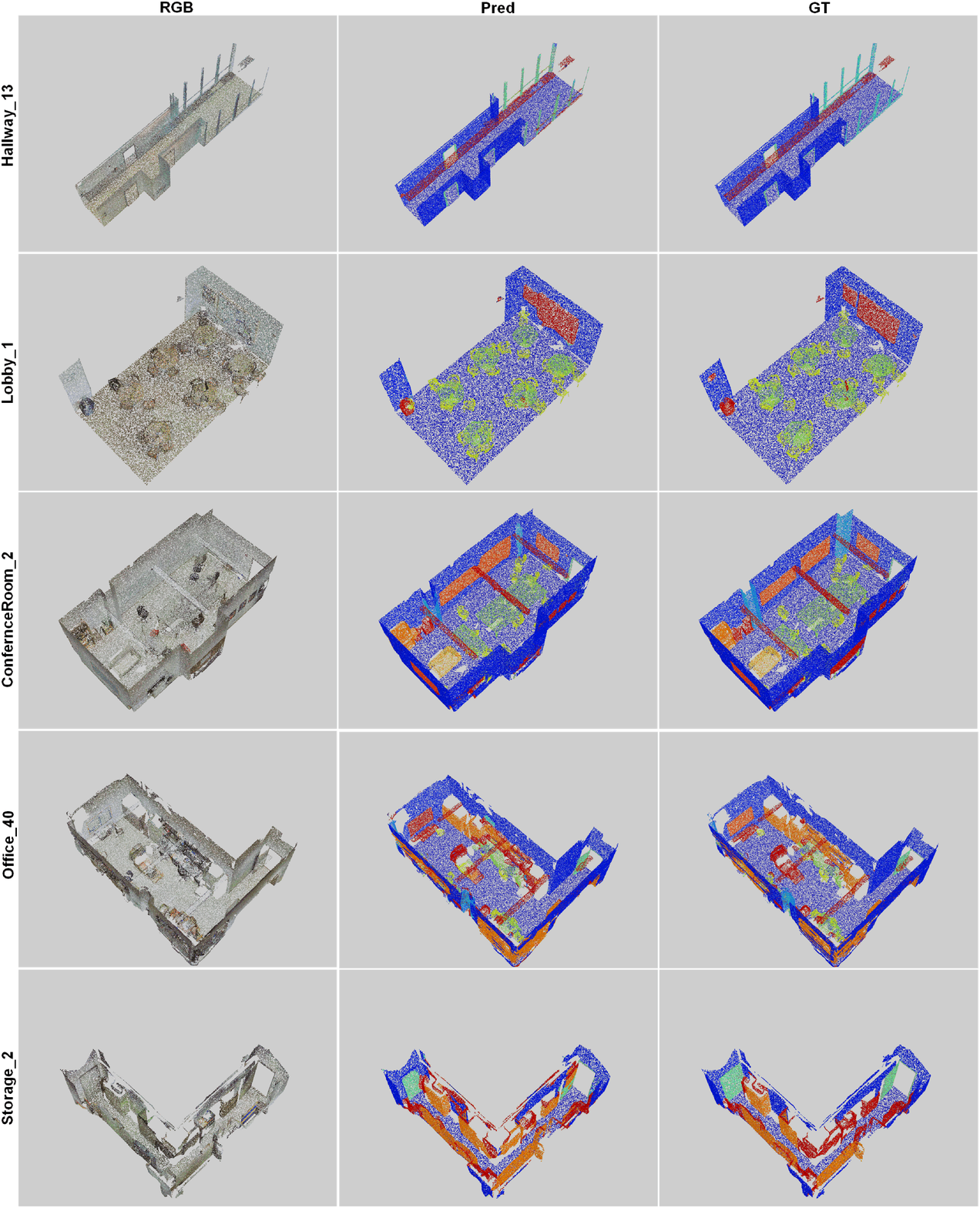}}
\caption{Visualization results compared with the groundtruth on Area 5 of the S3DIS dataset. Each class is visualized in a unique color. Note that the "ceiling" category is removed manually for better visualization.}
\label{fig2}
\end{figure}
\clearpage
\begin{figure}[!h]
	\centerline{\includegraphics[width=\columnwidth]{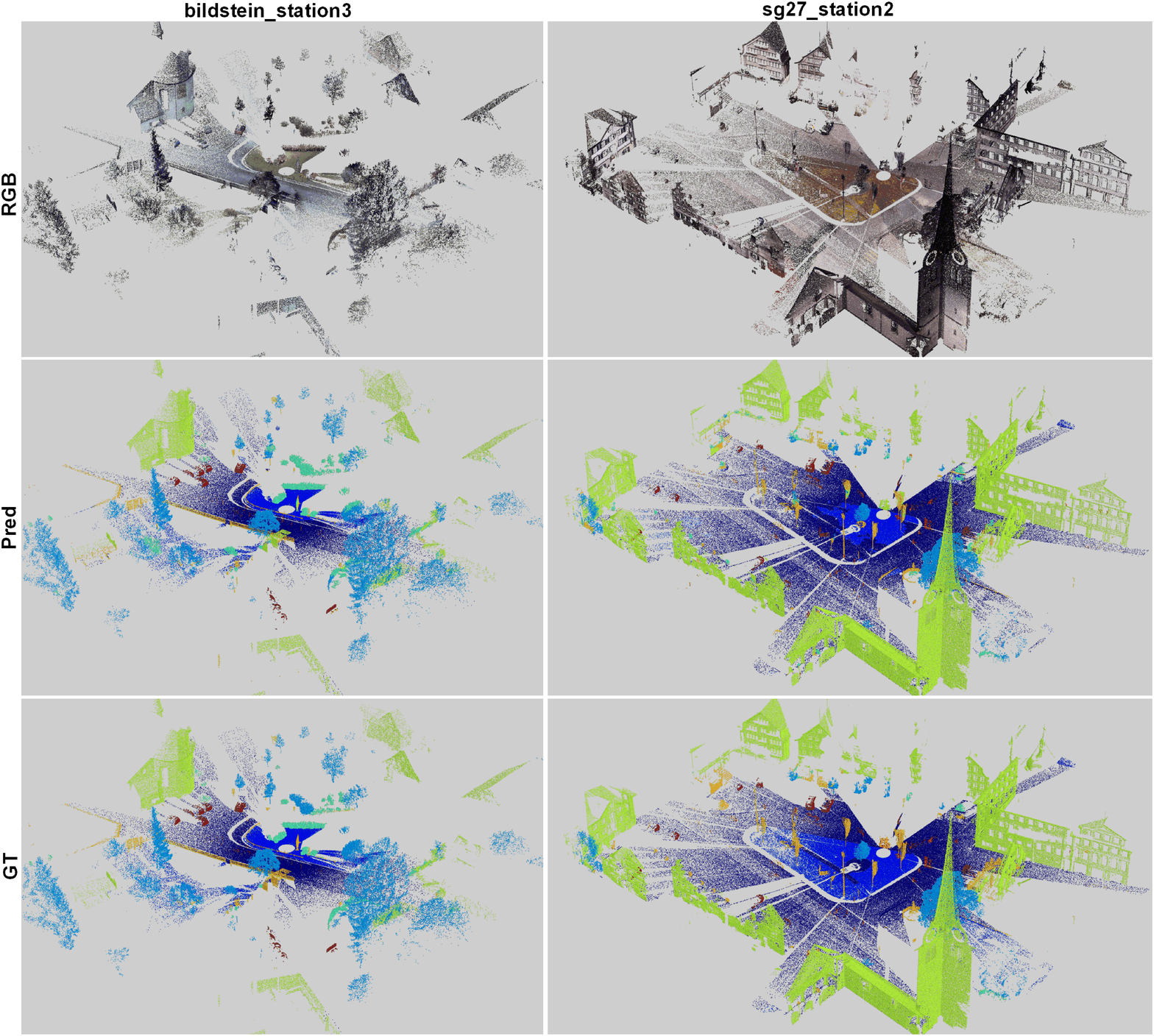}}
	\caption{Visualization results compared with the groundtruth on the validation set of the Semantic3D dataset. Each class is visualized in a unique color. Note that the "unlabeled" category is removed manually for better visualization.} 
	\label{fig3}
\end{figure}